%% file: main.tex
\documentclass[10pt]{article}
\usepackage[preprint]{tmlr}
\usepackage[english]{babel}

\usepackage{amsmath,amssymb,mathtools}
\usepackage{booktabs}
\usepackage{graphicx}
\usepackage{microtype}
\usepackage{xcolor}
\usepackage{float}
\usepackage{placeins}
\usepackage{url}
\usepackage{hyperref}
\usepackage{etoolbox}
\usepackage{multirow}
\usepackage{longtable}
\usepackage{array}
\usepackage{adjustbox}
\input{figure-titles.tex}
\AtBeginEnvironment{longtable}{\renewcommand\arraystretch{0.82}}
\definecolor{dmtblue}{HTML}{2563A6}
\definecolor{wipred}{HTML}{A61B1B}
\definecolor{wipbg}{HTML}{FFF3F3}

\newcounter{xaviercounter}

\newcounter{yanniscounter}

\hypersetup{colorlinks=true,linkcolor=dmtblue,citecolor=dmtblue,urlcolor=dmtblue}
\newcommand{\DMT}{\textsc{DMT}}

\newcommand{\R}{\mathbb{R}}
\newcommand{\C}{\mathbb{C}}
\newcommand{\bigO}{\mathcal{O}}
\newcommand{\xpfig}[4][width=\linewidth,height=0.63\textheight,keepaspectratio]{%
  \begin{figure}[H]
    \centering
    \begin{adjustbox}{max width=\linewidth,max totalheight=0.78\textheight}
    \includegraphics[#1]{fig_xp/#2.pdf}
    \end{adjustbox}
    \caption[\ThesisDMTFigureTitle{#2}]{#3}
    \label{fig:#4}
  \end{figure}}

\title{Dynamical Memory Transformer: Selective Recurrent Memories with Constant Cost Attention}

\author{\name Yannis Bendi-Ouis \email yannis.bendi-ouis@inria.fr \\
      \addr Inria Centre at the University of Bordeaux, Bordeaux, France \\\\
      \name Xavier Hinaut \email xavier.hinaut@inria.fr \\
      \addr Inria Centre at the University of Bordeaux, Bordeaux, France}

\makeatletter
\patchcmd{\@maketitle}{Anonymous authors\\Paper under double-blind review}{Anonymous authors\\Paper under double-blind review}{}{}
\makeatother
\renewenvironment{abstract}{\vskip.075in\centerline{\large\bf\sffamily Abstract}\vspace{0.5ex}\begin{quote}}{\par\end{quote}\vskip 1ex}
\begin{document}
\input{body.tex}
\input{appendices.tex}

\clearpage
\input{generations_owt.tex}
\end{document}

%% file: figure-titles.tex
\newcommand{\ThesisDMTFigureTitle}[1]{\csname ThesisDMTFigureTitle@#1\endcsname}
\expandafter\def\csname ThesisDMTFigureTitle@cogscale_results\endcsname{CogScale performance versus model size}
\expandafter\def\csname ThesisDMTFigureTitle@extra_context\endcsname{Selective copy beyond the training length}
\expandafter\def\csname ThesisDMTFigureTitle@activity_unit_per_input_word\endcsname{Memory activity by input word}
\expandafter\def\csname ThesisDMTFigureTitle@activity_aligned_units\endcsname{Information decoded from the memory units}
\expandafter\def\csname ThesisDMTFigureTitle@activity_unit_ablation\endcsname{Effect of removing individual memory outputs}
\expandafter\def\csname ThesisDMTFigureTitle@shakespeare_learning_curves\endcsname{Tiny Shakespeare learning curves}
\expandafter\def\csname ThesisDMTFigureTitle@shakespeare_epochs_vs_test_accuracy\endcsname{Tiny Shakespeare accuracy versus training epochs}
\expandafter\def\csname ThesisDMTFigureTitle@tinystories_stage1_feature_correlations\endcsname{TinyStories architectural parameter correlations}
\expandafter\def\csname ThesisDMTFigureTitle@tinystories_winners_validation_test\endcsname{TinyStories results for the selected configurations}
\expandafter\def\csname ThesisDMTFigureTitle@owt_training_results\endcsname{OpenWebText language modeling results}
\expandafter\def\csname ThesisDMTFigureTitle@owt_lm_eval_language\endcsname{Zero-shot evaluation on WikiText-2 and LAMBADA}
\expandafter\def\csname ThesisDMTFigureTitle@owt_lm_eval_choice\endcsname{Zero-shot multiple-choice evaluation}
\expandafter\def\csname ThesisDMTFigureTitle@compute_time\endcsname{Training and initial context processing time}
\expandafter\def\csname ThesisDMTFigureTitle@compute_memory\endcsname{Hardware memory during training and context processing}
\expandafter\def\csname ThesisDMTFigureTitle@compute_decode_state\endcsname{Autoregressive latency and retained state size}
\expandafter\def\csname ThesisDMTFigureTitle@ablation_results\endcsname{Effect of DMT component ablations}
\expandafter\def\csname ThesisDMTFigureTitle@ablation_score_vs_training_time\endcsname{Ablation performance versus training time}
\expandafter\def\csname ThesisDMTFigureTitle@cogscale_dmt_hp_correlation\endcsname{DMT parameter correlations on CogScale}
\expandafter\def\csname ThesisDMTFigureTitle@activity_real_sentences\endcsname{Memory activity over Cross Situation sentences}
\expandafter\def\csname ThesisDMTFigureTitle@activity_similarity_attention\endcsname{Memory similarity and decoder attention}
\expandafter\def\csname ThesisDMTFigureTitle@tinystories_hp_lr_weight_decay\endcsname{TinyStories learning rate and weight decay search}
\expandafter\def\csname ThesisDMTFigureTitle@tinystories_hp_best_per_seed\endcsname{TinyStories test perplexity across seeds}
\expandafter\def\csname ThesisDMTFigureTitle@tinystories_warmup_floor\endcsname{TinyStories warmup and learning rate floor search}
\expandafter\def\csname ThesisDMTFigureTitle@tinystories_config_stage1_top15\endcsname{TinyStories first-stage configuration selection}
\expandafter\def\csname ThesisDMTFigureTitle@tinystories_config_stage2_top5\endcsname{TinyStories second-stage configuration selection}
\expandafter\def\csname ThesisDMTFigureTitle@tinystories_config_stage3_validation_ranking\endcsname{TinyStories final validation ranking}
\expandafter\def\csname ThesisDMTFigureTitle@tinystories_config_stage3_test_descriptive\endcsname{TinyStories test results for the final configurations}
\expandafter\def\csname ThesisDMTFigureTitle@owt_learning_curves\endcsname{OpenWebText learning curves}

%% file: body.tex
\raggedbottom
\maketitle

\begin{abstract}
Transformers retain an explicit history of the sequence, allowing attention to directly retrieve past representations. However, the size of the KV-Cache and the cost of a new token increase with the context length. State Space Models compress the past into a fixed-size recurrent state, at the cost of reduced addressability. We present the Dynamical Memory Transformer (\DMT), a hybrid architecture that separates sequence encoding from information reading. The encoder uses several memory units, each implemented by a dynamical system with a complex diagonal recurrent matrix. An input-dependent leak rate controls their updates. The decoder then uses attention to read and combine their outputs, without going through the token history. The formulation of the recurrence allows an associative scan with $\bigO(T)$ total work and $\bigO(\log T)$ parallel depth, with fixed dimensions. During autoregressive inference, the recurrent states and the convolution cache remain fixed in size. We compare DMT with Mamba and Transformer on cognitive and language modeling tasks, and show its strengths and limitations: 
DMT achieves the best results on several cognitive tasks and language modeling performance close to that of Transformer, but remains behind Mamba in language modeling. On selective copy, DMT also shows stronger length extrapolation than Mamba and Transformer. These results demonstrate that parallel dynamical memory units are a promising mechanism for retaining and using information over sequences.
\end{abstract}

\section{Introduction}

Sequential processing requires selecting, maintaining and using, at the right time, the information contained in past inputs. Transformer architectures retain an explicit history of the representations of each token for this purpose~\citep{vaswani2017attention}. At each step, attention can retrieve a past representation from its content, without going through all the updates that separate it from the current token. We call this possibility the addressability of information. During training, known tokens can be processed in parallel. However, during autoregressive inference, the KV-Cache that retains keys and values grows with the context. The total work of dense attention is quadratic in sequence length, and the cost of each new token grows linearly. FlashAttention reduces memory transfer volume and improves GPU utilization~\citep{dao2022flashattention}, without eliminating this growth. \\

\noindent Recurrent architectures follow another strategy: they progressively compress the past into a fixed-size recurrent state. This state evolves with each input, but its dimension does not depend on the sequence length. It therefore allows constant incremental complexity during inference, for a given architecture. In return, information that has not been retained in the state can no longer be retrieved directly. State Space Models (SSMs) combine this principle with parallel formulations: S4 notably uses a convolutional view of the recurrence~\citep{gu2022s4}, while Mamba allows input-dependent selection and uses an associative scan~\citep{gu2023mamba}. These two strategies therefore contrast an explicit addressable history with compression of the past into a fixed-size state. \\

\noindent Here, we explore a separation between encoding and reading. The Dynamical Memory Transformer (\DMT) uses $M$ memory units, each with its own recurrent parameters. At each step, an input-dependent leak rate controls the interpolation between the previous state and a candidate state. Together, these units serve as an encoder of the sequence. The decoder then uses attention to read and combine the resulting representations, rather than going through past tokens. In a functional sense, these states constitute working memory: a temporary representation of information useful for the current sequence. \\

\noindent Our contributions are as follows:
\begin{itemize}
  \item a hybrid architecture that separates an encoder with selective memory units from an attention decoder, with a fixed-size memory medium during inference;
  \item an exact formulation of the recurrence through associative scan, which distinguishes total work, the parallel depth of the algorithm and the limitations of the chunked implementation;
  \item a comparison on cognitive tasks, memory unit activity, length extrapolation and language modeling;
  \item a fixed-budget ablation of the selected TinyStories configuration, to measure the role of the components and their interactions.
\end{itemize}

\noindent The experiments show that this separation allows several tasks to be learned, but also that compression can limit recall. 

\section{Related work}

\paragraph{Attention and cost reduction.}
The attention mechanism of Transformers reads a history of representations using queries, keys and values~\citep{vaswani2017attention}. FlashAttention computes this same attention while reducing transfers between the GPU's HBM and its smaller, faster SRAM~\citep{dao2022flashattention}. Kernel-based linear attention replaces explicit history with associative compression, which admits a recurrent form~\citep{katharopoulos2020linear}. RetNet also offers parallel, recurrent and chunkwise recurrent forms~\citep{sun2023retnet}. In \DMT, the difference concerns the medium read by attention: it consults the outputs of the memory units at the current time.

\paragraph{Recurrent architectures and State Space Models.}
RNNs maintain a recurrent state that summarizes past inputs. Reservoir Computing offers another training method: recurrent parameters remain unchanged and a readout is learned to read the states~\citep{jaeger2001echo,maass2002liquid}. The eigenvalues of the recurrent matrix and the leak rate then influence the dynamical regime of the Reservoir. S4 uses a structured state-space system whose recurrent and convolutional forms allow computation to be adapted to inference and training~\citep{gu2022s4}. H3 constructs and memorizes key--value associations using two SSMs~\citep{fu2023h3}. Mamba makes some parameters input-dependent, allowing the writing and maintenance of information to be modulated~\citep{gu2023mamba}. Mamba-2 connects structured SSMs and attention through a duality between their formulations~\citep{dao2024mamba2}. RWKV also retains a recurrent form for inference and a parallel form for training~\citep{peng2023rwkv}. \DMT{} distributes compression across several memory units, then assigns their reading to a decoder.

\paragraph{Evaluation at several scales.}
CogScale provides synthetic cognitive tasks whose difficulty, amount of data and parameter budget can vary~\citep{anonymous2026cogscale}. Tiny Shakespeare allows small-scale next-character prediction to be tested~\citep{karpathy2015charrnn}. TinyStories provides a corpus on which small architectures can learn token prediction and produce coherent text~\citep{eldan2023tinystories}. Finally, OpenWebText allows comparison on a larger and more diverse corpus~\citep{gokaslan2019openwebtext}. We complement these evaluations by studying unit activity and the transfer of a copying rule to longer sequences.

\clearpage
\section{Dynamical Memory Transformer}
\label{sec:method}

\subsection{Overview and notation}

\begin{figure}[H]
  \centering
  \includegraphics[width=0.9\linewidth]{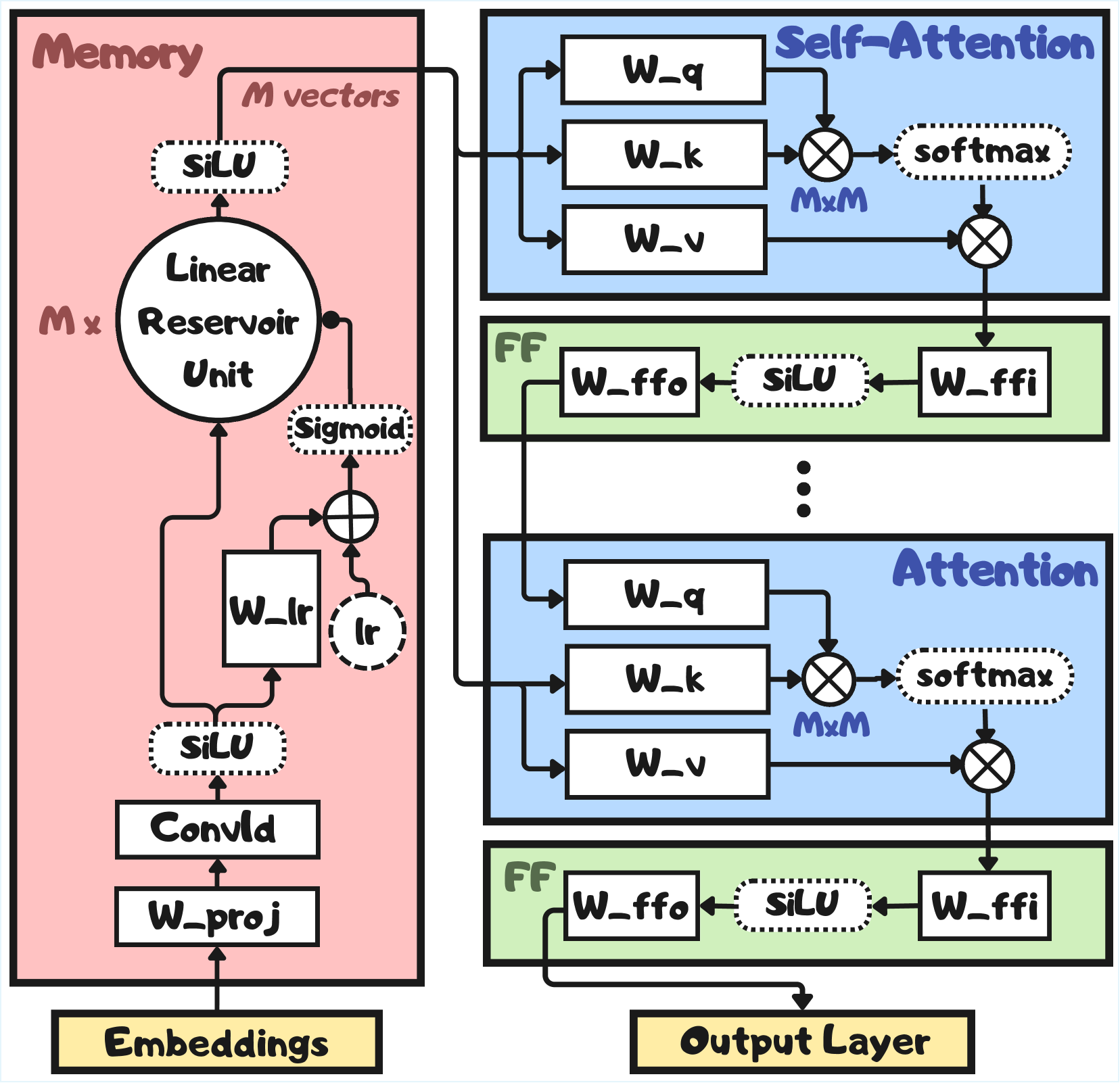}
  \caption[DMT architecture]{Overview of \DMT. The memory units progressively encode the sequence into fixed-size recurrent states. Decoder attention reads and combines their outputs at the current time. Between two inference steps, the architecture retains the recurrent states and the convolution cache, without retaining an explicit history of all tokens.}
  \label{fig:architecture}
\end{figure}

Let $x_t$ be the input vector associated with the current token, $D$ the main dimension of the architecture, $M$ the number of memory units, $N$ the dimension of the recurrent state per unit, $K$ the convolution filter width and $L$ the number of decoder layers. Each unit $m$ receives its own projection of $x_t$, applies a temporal 1D convolution over the last $K$ tokens, then a SiLU activation function. It then updates its complex recurrent state $h_{m,t}\in\C^N$ and produces a readout in $\R^D$. Together, the units form the encoder of the sequence. The decoder uses attention and feedforward layers to read and combine the $M$ representations obtained at each time. Figure~\ref{fig:architecture} presents this separation.

\FloatBarrier

\subsection{Selective memory units}

Each unit has an independent input projection $W_{\mathrm{proj},m}\in\R^{D\times D}$. The projection, RMS normalization and temporal convolution prepare the input used to update its state:
\begin{align}
  z_{m,t} &= \operatorname{SiLU}\!\left(\operatorname{Conv}_K(W_{\mathrm{proj},m}x_{1:t})_t\right) \\
  \alpha_{m,t} &= \sigma\!\left(W_{\alpha,m}z_{m,t} + b_{\alpha,m}\right)\in(0,1)^N 
\end{align}
The leak rate $\alpha_{m,t}$ depends on the input and is computed for each element of the state. The learned vector $\lambda_m\in\C^N$ contains the diagonal entries of the recurrent matrix $W_{\mathrm{rec},m}=\operatorname{diag}(\lambda_m)\in\C^{N\times N}$. Its components are initialized on the unit circle, then their real and imaginary parts are trained without additional constraints. The update coefficients are
\begin{align}
  a_{m,t} &= (1-\alpha_{m,t}) + \alpha_{m,t}\odot\lambda_m \\
  b_{m,t} &= \alpha_{m,t}\odot\left(W_{\mathrm{in},m}z_{m,t}\right)
\end{align}
where $W_{\mathrm{in},m}\in\C^{N\times D}$. The recurrence is written as:
\begin{equation}
  h_{m,t} = a_{m,t}\odot h_{m,t-1} + b_{m,t}
  \label{eq:recurrence}
\end{equation}
This update corresponds to an interpolation between the previous state and a candidate state that combines the new input and the transformation of the previous state by the recurrent matrix. When $\alpha\approx 0$, the state is almost preserved; for $\alpha=0$, it would be preserved exactly. When $\alpha\approx 1$, the system mainly applies the candidate update. The leak rate controls maintenance and writing of information according to the input token.

Reading the state of each unit produces the output:
\begin{equation}
  s_{m,t} = x_t + \operatorname{SiLU}\!\left(\Re\{W_{\mathrm{out},m}h_{m,t}\}\right)
  \qquad W_{\mathrm{out},m}\in\C^{D\times N}
\end{equation}
The complex input and output projections are learned. The residual connection directly adds $x_t$ to this readout of the state. The dependence on the previous state is linear, and the input adds a term at each step. However, the activation functions and the input-dependent leak rate make the complete architecture nonlinear.

\subsection{Associative scan}

Equation~\ref{eq:recurrence} allows online computation: the current state is obtained from the previous state and the current input. During training, however, the entire input sequence is known. To avoid a chain of $T$ sequential updates, we represent each transition $f_t(h)=a_t\odot h+b_t$ by a pair $(a_t,b_t)$. Two successive steps can then be grouped into a new pair of the same size:
\begin{equation}
  (a_1,b_1)\otimes(a_2,b_2)
  = \left(a_2\odot a_1,\; a_2\odot b_1+b_2\right)
  \label{eq:operator}
\end{equation}
This operator corresponds to the composition of transitions and is associative. Their grouping can therefore be changed without changing the result. For three steps, both parenthesizations give
\begin{equation}
  \left(a_3a_2a_1,\;a_3a_2b_1+a_3b_2+b_3\right)
\end{equation}
where products are performed elementwise. However, temporal order must be preserved: the operator is generally not commutative. An associative prefix scan~\citep{blelloch1990prefix} allows transitions to be grouped into a tree and all states $h_t$ to be computed. With fixed dimensions, total work remains $\bigO(T)$, while parallel depth is $\bigO(\log T)$ for the algorithm over the complete sequence. These two complexities measure different properties: the second describes necessarily sequential steps when resources allow maximum parallelism.

The implementation uses a custom Triton kernel~\citep{tillet2019triton}. The scan is performed within chunks of length $C$, then a sequential host-side loop passes the state from one chunk to the next. The parallel depth of this implementation is therefore $\bigO(\!\lceil T/C\rceil+\log C)$; with a single chunk, it returns to $\bigO(\log T)$. A hierarchical scan entirely executed on the GPU remains to be implemented. Temporal parallelism applies to training and initial context processing. During autoregressive inference, each new token must still be produced before the next.

\subsection{Attention across memories}

At time $t$, the outputs of the memory units are grouped into $S_t\in\R^{M\times D}$. The first decoder block applies self-attention to these $M$ representations. For subsequent blocks, queries come from the output of the previous block, while keys and values still come from the encoder outputs. Each block includes attention with a residual connection, RMS normalization and a feedforward layer. After $L$ blocks, the $M$ representations are concatenated, projected into dimension $D$, then transformed into logits to predict the next token.

Attention thus allows the information retained by the different units to be read and combined. Its addressability concerns these representations, not the complete token history. If information has been erased from all states, the decoder can no longer retrieve it. The capacity of the encoder and the selection performed during updates therefore determine what the decoder can read.

\subsection{Complexity}

Here, we distinguish total work, parallel depth and the size of the memory medium retained during inference. Incremental complexity measures the cost of processing a new token. For \DMT, it can be approximated by
\begin{equation}
  \bigO\!\left(MD^2+MDN+L(MD^2+M^2D)\right)
  \label{eq:work}
\end{equation}
where the terms correspond to the projections of each unit, writing and reading of the recurrent states, the projections and feedforward layers of the decoder, then attention across units. The dimensions $M$, $D$, $N$, $K$ and the number of layers $L$ are fixed when the architecture is created. Incremental complexity therefore does not depend on the context length; total work over $T$ tokens is $T$ times this expression. However, the cost per token can remain high when these dimensions increase.

The memory medium passed between two steps contains the recurrent states and the $K-1$ previous convolution inputs, or $\bigO(MN+MD(K-1))$ values. Its size is independent of $T$. This does not mean that all hardware memory used during training is constant: the activation memory needed to compute the gradient also depends on the sequence length. Finally, execution time depends on the hardware, memory transfer volume and kernel quality, beyond these asymptotic expressions.

\begin{table}[H]
\centering
\small
\label{tab:complexity}
\begin{tabular}{lccc}
  \toprule
  Family & Total work & Parallel depth & Inference memory medium \\
  \midrule
  Dense Transformer & $\bigO(LT^2D)$ & $\bigO(L)$ layers & $\bigO(LTD)$ \\
  Selective diagonal SSM & $\bigO(TDN)$ & $\bigO(\log T)$, ideal scan & $\bigO(N)$ \\
  \DMT{} & $T\times$ Equation~\ref{eq:work} & $\bigO(\log T)+\bigO(L)$, ideal & $\bigO(MN+MDK)$ \\
  \bottomrule
\end{tabular}
\caption{Asymptotic comparison for a sequence length $T$. The table distinguishes total work during training, parallel depth and the memory medium retained between two inference steps. Constants, the vocabulary projection, batch size and memory transfers are omitted. The Transformer and SSM rows give simplified costs of attention or state processing; projections and feedforward layers are not all detailed.}
\end{table}


\section{Experiments and results}
\label{sec:experiments}
\label{sec:results}

\subsection{Overview of the experiments}

To evaluate the Dynamical Memory Transformer as thoroughly as possible, we performed several distinct experiments to answer the following questions: 
(i) With a limited parameter budget, can DMT learn and scale on different cognitive tasks?
(ii) Is it possible to transfer the rules learned by DMT beyond the sequence lengths seen during training? 
(iii) What do DMT's recurrent units encode, and do they learn to specialize? 
(iv) How does DMT behave in character-level and token-level language modeling, and at different scales? 
(v) Which components contribute to the performance of the selected configuration? 
Except for question (iii), which is specific to DMT, we systematically compared DMT with Transformer and Mamba. For this purpose, we focused on 4 datasets. CogScale~\citep{anonymous2026cogscale} provides a set of synthetic, scalable cognitive tasks that notably allowed us to iterate quickly at a small scale while building the model and conducting its first tests. Here, it will be useful for comparing small-scale model performance and activity. To evaluate DMT on text modeling, we selected the Tiny Shakespeare~\citep{karpathy2015charrnn}, TinyStories~\citep{eldan2023tinystories} and OpenWebText~\citep{gokaslan2019openwebtext} text corpora, with respective sizes of 1 MB, 2 GB and 40 GB. These different scales will allow us to study models of different sizes: 300k, 10M and 125M parameters.

\subsection{CogScale}

\paragraph{Experimental protocol.}
The two CogScale figures show results for \texttt{bracket\_matching}, \texttt{induction\_heads}, \texttt{selective\_copy}, \texttt{simple\_copy} and \texttt{sorting\_problem}, at the \texttt{small} and \texttt{medium} difficulty levels. \texttt{bracket\_matching} asks whether a bracket sequence is valid. \texttt{induction\_heads} asks for the value associated with a key when that key appears a second time. \texttt{simple\_copy} consists of reproducing a sequence after a delay, while \texttt{selective\_copy} asks for only the marked symbols to be reproduced. Finally, \texttt{sorting\_problem} asks for symbols to be reordered according to the position indices provided in the input.
The budgets studied are 1k, 10k and 100k parameters. Since we could not know which architecture would favor which configuration with which hyperparameters, we performed a grid search. Thus, for each architecture, each budget, each task and each difficulty, we train 500 models: 10 configurations, 5 learning rates, 2 weight decays and 5 seeds ranging from 0 to 4. This gives a total of 45,000 models. Training uses Adam, batches of 10, a maximum of 200 epochs and a patience of 10 epochs. 
The five tasks evaluate different operations on sequences.

\begin{table*}[h]
  \centering
  \small
  \setlength{\tabcolsep}{3pt}
  \renewcommand{\arraystretch}{1.08}
  \begin{tabular*}{\textwidth}{@{\extracolsep{\fill}}lllr@{\hspace{1.5em}}lllr@{}}
    \toprule
    Task & Difficulty & Architecture & Accuracy (\%) & Task & Difficulty & Architecture & Accuracy (\%) \\
    \cmidrule(lr){1-4}\cmidrule(lr){5-8}
    \multirow{6}{*}{\shortstack[l]{Bracket\\matching}} & \multirow{3}{*}{Small} & DMT & $\mathbf{88.20 \pm 9.55}$ & \multirow{6}{*}{\shortstack[l]{Induction\\heads}} & \multirow{3}{*}{Small} & DMT & $99.98 \pm 0.05$ \\
    &  & Mamba & $72.40 \pm 12.20$ &  &  & Mamba & $53.26 \pm 5.29$ \\
    &  & Transformer & $60.20 \pm 10.80$ &  &  & Transformer & $\mathbf{100.00 \pm 0.00}$ \\
    \cmidrule(lr){2-4}\cmidrule(lr){6-8}
    & \multirow{3}{*}{Medium} & DMT & $98.14 \pm 0.15$ &  & \multirow{3}{*}{Medium} & DMT & $99.87 \pm 0.29$ \\
    &  & Mamba & $\mathbf{98.44 \pm 0.71}$ &  &  & Mamba & $90.12 \pm 4.39$ \\
    &  & Transformer & $93.96 \pm 3.41$ &  &  & Transformer & $\mathbf{100.00 \pm 0.00}$ \\
    \cmidrule(lr){1-4}\cmidrule(lr){5-8}
    \multirow{6}{*}{\shortstack[l]{Selective\\copy}} & \multirow{3}{*}{Small} & DMT & $\mathbf{57.80 \pm 28.37}$ & \multirow{6}{*}{\shortstack[l]{Simple\\copy}} & \multirow{3}{*}{Small} & DMT & $99.82 \pm 0.26$ \\
    &  & Mamba & $57.56 \pm 13.87$ &  &  & Mamba & $50.99 \pm 0.80$ \\
    &  & Transformer & $39.96 \pm 15.12$ &  &  & Transformer & $\mathbf{100.00 \pm 0.00}$ \\
    \cmidrule(lr){2-4}\cmidrule(lr){6-8}
    & \multirow{3}{*}{Medium} & DMT & $\mathbf{98.68 \pm 2.40}$ &  & \multirow{3}{*}{Medium} & DMT & $99.76 \pm 0.39$ \\
    &  & Mamba & $93.53 \pm 6.17$ &  &  & Mamba & $87.89 \pm 10.18$ \\
    &  & Transformer & $85.82 \pm 14.51$ &  &  & Transformer & $\mathbf{100.00 \pm 0.00}$ \\
    \cmidrule(lr){1-4}\cmidrule(lr){5-8}
    &  &  &  & \multirow{6}{*}{\shortstack[l]{Sorting\\problem}} & \multirow{3}{*}{Small} & DMT & $48.26 \pm 1.59$ \\
    &  &  &  &  &  & Mamba & $61.24 \pm 5.51$ \\
    &  &  &  &  &  & Transformer & $\mathbf{100.00 \pm 0.00}$ \\
    \cmidrule(lr){6-8}
    &  &  &  &  & \multirow{3}{*}{Medium} & DMT & $62.03 \pm 5.55$ \\
    &  &  &  &  &  & Mamba & $95.55 \pm 5.40$ \\
    &  &  &  &  &  & Transformer & $\mathbf{100.00 \pm 0.00}$ \\
    \bottomrule
  \end{tabular*}
  \caption{Results averaged over 5 seeds for the best configuration obtained through grid search. The reported score is mean accuracy in \% and its standard deviation. The best score for each task, difficulty and architecture is shown in bold.}
  \label{tab:cogscale-best}
\end{table*}

\paragraph{Results.}

Table~\ref{tab:cogscale-best} presents, for each task,
difficulty, architecture, budget and hyperparameter, the configuration  
achieving the best mean accuracy 
over five seeds. On tasks such as \texttt{induction\_heads} and
\texttt{simple\_copy}, DMT achieves performance close to that of Transformer 
with accuracy above 99\% at
both difficulty levels. For \texttt{selective\_copy},
it achieves the best mean, particularly at \texttt{medium} difficulty,
with $98.68\%$, compared with $93.53\%$ for Mamba and
$85.82\%$ for Transformer. For the 
\texttt{bracket\_matching} task, DMT outperforms both other architectures
at \texttt{small} difficulty and is very slightly outperformed by Mamba at
\texttt{medium} difficulty: $98.14\%$ versus $98.44\%$. 
However, for \texttt{sorting\_problem}, DMT is consistently behind the
other two architectures. Apart from this last task, DMT performs very well on a range of cognitive tasks. At a small budget, it regularly ranks behind Transformer and its explicit addressable history, and very often ahead of Mamba. It even sometimes takes first place, as is the case for \texttt{selective\_copy}. 


Figure \ref{fig:xp-cogscale-results} presents, for each task, difficulty and budget, the mean accuracy obtained across all configurations and seeds for the best learning rate/weight decay combination for each architecture. These results therefore do not select an optimal configuration and show a mean result that could be obtained without an extensive hyperparameter search. 
We can observe that, in general, DMT improves as its number of parameters increases, validating its ability to scale at least under a small budget. We can note that it performs very well when transporting specific information over a long period of time, as with \texttt{selective\_copy}, while it is inferior to Transformer on tasks requiring an explicit addressable history, such as \texttt{induction\_heads}, \texttt{simple\_copy} and \texttt{sorting\_problem}. We can also note that, under a small budget, DMT is often superior or close to Mamba, except on \texttt{sorting\_problem}.

\xpfig{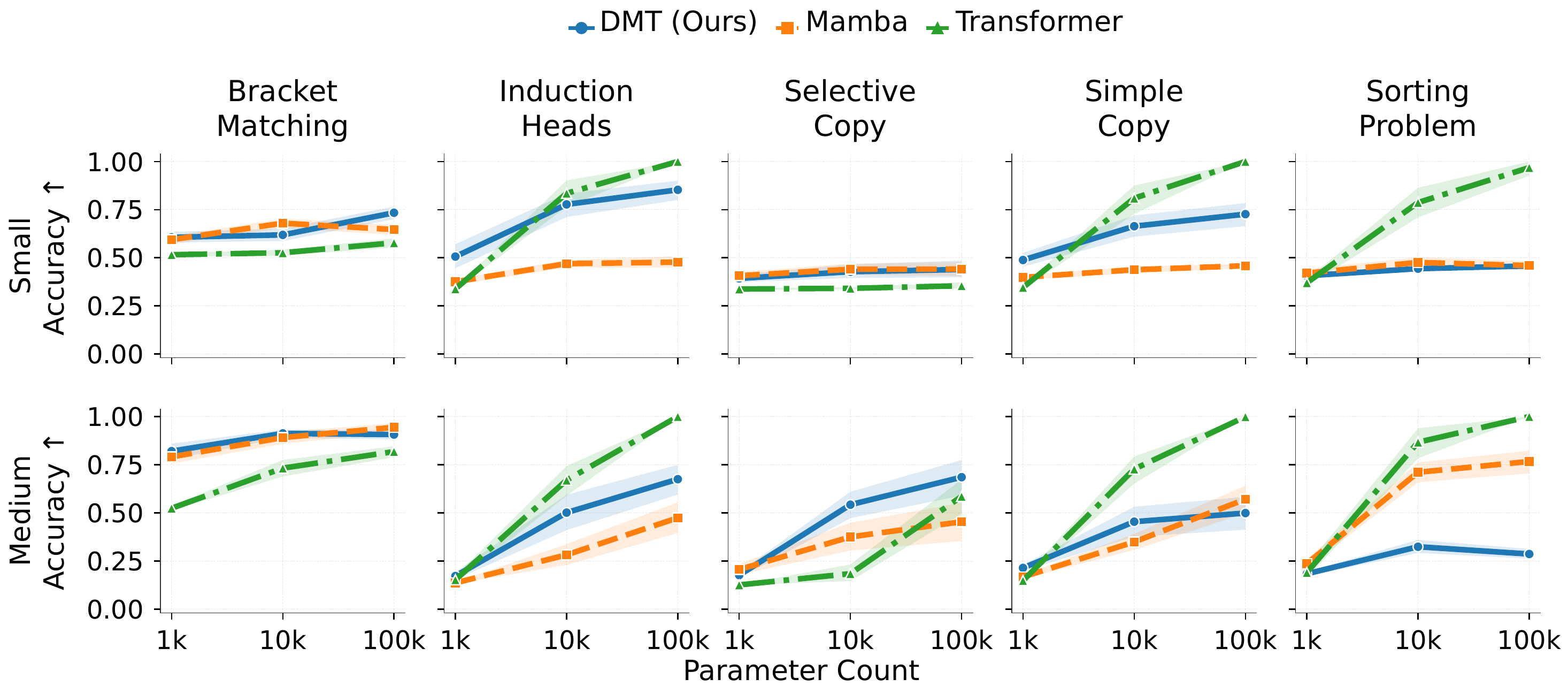}{Evolution of mean scores as a function of the number of parameters. These curves illustrate the ability of each architecture to scale under small budgets. Each point was obtained by averaging the results of 50 models, with the learning rate and weight decay fixed.}{xp-cogscale-results}


Having trained 45,000 models of different sizes and configurations, we sought to understand which DMT parameters had an influence on each task. For this purpose, we computed Spearman correlations between these different architectural parameters and the accuracies obtained on CogScale. Figure \ref{fig:xp-cogscale-hp} (Appendix~\ref{app:results-cogscale}) presents each of these scores. We can observe that, at least under a small budget, the number of decoder layers $L$ and the convolution filter length $K$ do not seem particularly important, except for \texttt{selective\_copy}, where the correlations suggest that a shorter convolution filter is preferable (since the symbols to reproduce are scattered among distractors, their immediate neighborhood provides little useful information). 
In contrast, the main model dimension $D$ and the number of decoder heads $n_{\mathrm{heads}}$ seem to play a major role in almost all tasks. Tasks requiring a large memory over a long duration, such as \texttt{induction\_heads} and \texttt{simple\_copy}, benefit substantially from increasing the number of memory units $M$ and their dimension $N$.


\subsection{Length extrapolation on selective copying}

\paragraph{Experimental protocol.}
After studying performance on CogScale, we sought to determine whether the learned dynamics could be used on sequences longer than those seen during training. For this purpose, we selected the \texttt{selective\_copy} task, which requires retaining ten relevant symbols within a sequence, then reproducing them. We train each architecture with approximately 100k parameters on five seeds ranging from 0 to 4, for a total of 15 models. Training sequences have a length of 256, corresponding to 277 tokens once control and response symbols are added. Each training run uses Adam for 25\,000 steps, with a new batch of 64 examples generated at each step. We then evaluate these same models, without further training, on sequences with lengths ranging from 64 to 262\,144. For each example, we measure the error on the ten tokens to predict and exact-match accuracy, which equals 1 only if all ten answers are correct. For sequences of length 2048 or less, the number of test examples per seed is 2\,048. Beyond this length, the number of examples progressively decreases to 16. To process these sequences, DMT and Mamba traverse the entire sequence while maintaining their recurrent state. Transformer, in contrast, has a sliding context window of 277 tokens, allowing it to see only the end of the sequence when making its prediction.

\paragraph{Results.}
Figure~\ref{fig:xp-extra-context} presents the evolution of performance as length increases. When the test length matches the training length, median performance is close to 100\% for all models. DMT and Mamba then maintain a median token-level error below 1\% up to length 1\,024. Beyond this, DMT maintains a consistently lower error than Mamba until the two meet at length 65,536. 
We can therefore observe that, on this task, DMT manages to retain information over a greater distance. However, as length increases, both architectures eventually fail. 
Transformer, in contrast, already reaches a median error of 78.48\% at length 512, with zero median exact-match accuracy. The symbols to copy progressively leave its context window, whereas DMT and Mamba can retain their information in their recurrent state. 

\xpfig{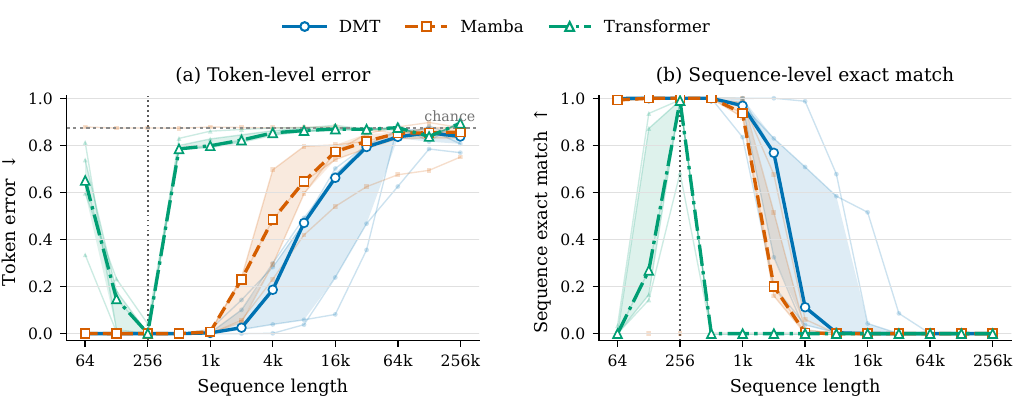}{Evolution of performance on \texttt{selective\_copy} as sequence length increases, after training at length 256. Thick curves show the median over five seeds, shaded areas the interquartile range and thin curves each seed. DMT and Mamba process the entire sequence; Transformer uses a sliding window of 277 tokens.}{xp-extra-context}


\subsection{Memory activity on Cross Situation}

\paragraph{Experimental protocol.}
We then sought to understand what the memory units encode and whether they learn to specialize. For this purpose, we use the CogScale \texttt{cross\_situation} task at \texttt{medium} difficulty. Each sample contains a sentence that successively describes two situations, and the model must perform multi-label classification over 28 binary classes indicating the presence of an object, its color and its position. Some words can be ambiguous, as is the case for "orange", which can be an object or a color. Two different words can indicate the same class to predict, as is the case for "centre" and "milieu", which indicate the same position. Each of these words is labeled according to its role in the sentence as \textit{Objects}, \textit{Colors}, \textit{Positions}, \textit{Relations} or \textit{Grammar}.

To conduct this experiment, we trained the same DMT configuration with 4 memory units and 111,312 parameters on 5 seeds (0 to 4). Each training run uses 1000 samples, Adam as the optimizer, binary cross-entropy as the loss, and validation-based early stopping. The test set also contains 1\,000 samples. For each input word, we record the leak rate, the recurrent state, its variation, the output of each memory unit and the decoder attention weights. The leak rate is particularly useful, as it indicates how much the unit updates. That said, it is not sufficient to measure the amount of information written. We therefore also measure state variation with:

$$
\delta_{m,t}=\frac{1}{N}\sum_n|h_{m,t,n}-h_{m,t-1,n}|
$$

We also sought to determine which information contained in these memory units was linearly decodable. For this purpose, we attached a linear readout to the final state of each unit, whose role is to recover the properties of the sentence. They are trained on the training examples and then evaluated on the test set. 
Since memory units are agnostic to their position, their roles can switch from one seed to another. For this reason, we aligned the units based on their prediction profiles. 
Finally, we sought to measure the importance of each memory unit for the model and their redundancy. For this purpose, we ablated these units, compared the similarity of their final states and observed the mean attention score assigned by the decoder to each unit.

\paragraph{Results.}
The five models achieve a mean exact-match accuracy of $99.08\%\pm0.70\%$ on the test set. DMT therefore correctly learns the \texttt{cross\_situation} task with four memory units.


Figure~\ref{fig:xp-activity-word} shows the mean leak rate of each unit and its mean variation for each word.
We can observe that unit \texttt{U1} is the only one interested in \textit{Positions}. Objects and colors are more shared across the other units, although we can note that \texttt{U4} seems to specialize more in \textit{Relations} and \textit{Grammar}, and \texttt{U3} in \textit{Objects} and \textit{Relations}. \texttt{U2} seems more general than the others. We can also notice that the special word "and", present in all samples and separating the two situations, is the only word outside \textit{Objects} and \textit{Relations} captured by \texttt{U3}. 
Comparing these two measures, it does indeed seem that a high leak rate leads to a large state variation, although this is more pronounced when the unit is highly specialized, as is the case for \texttt{U1}. This can be explained by the fact that \texttt{U1} receives almost no information except when a position appears in the input. Its state therefore has a low value most of the time, then suddenly changes when a position arrives.

\xpfig{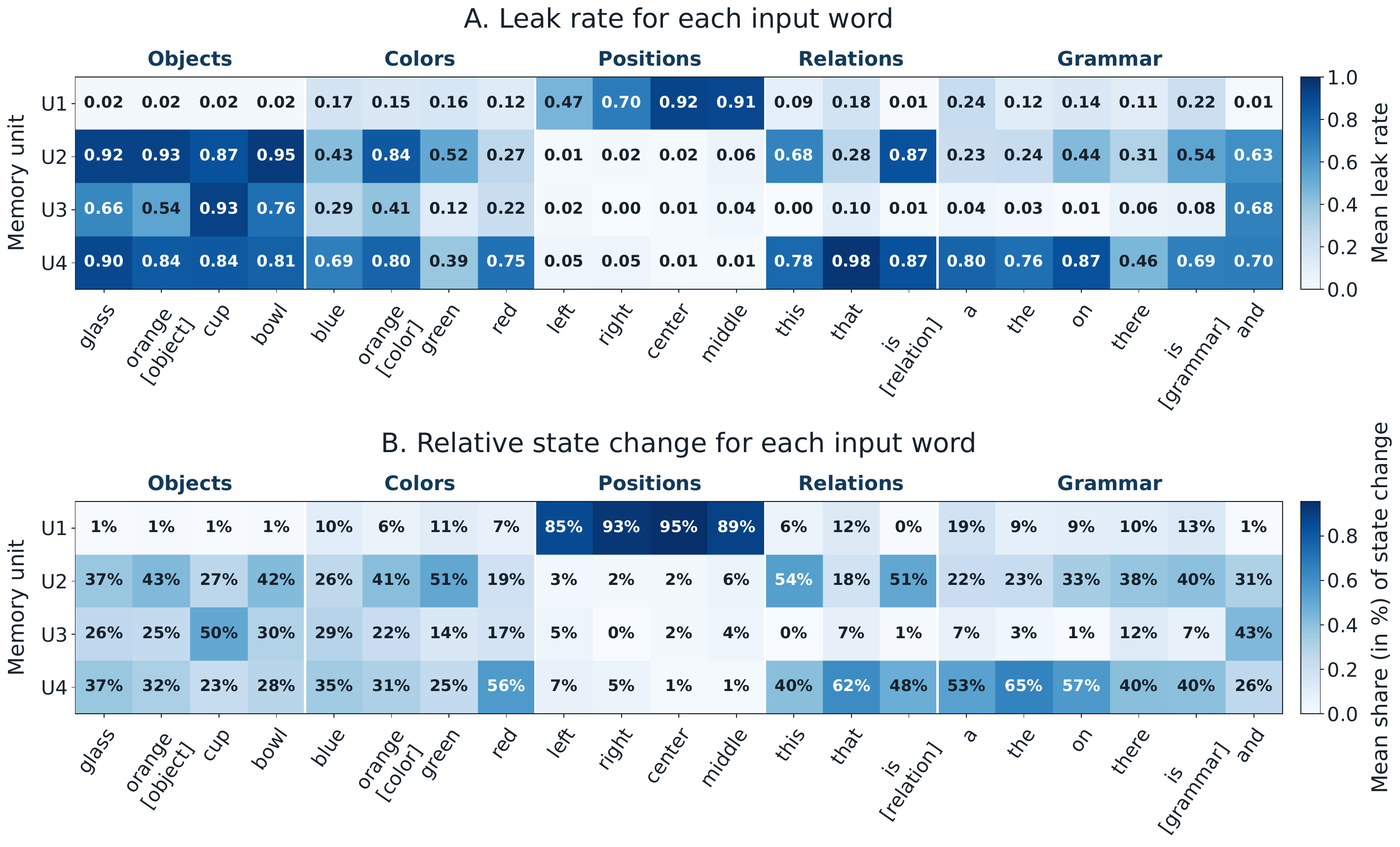}{Activity of the four memory units as a function of the words in \texttt{cross\_situation}. Plot A shows the mean leak rate, and plot B the mean state variation. The words at the bottom of each plot correspond to the vocabulary for this task. The categories at the top of each plot correspond to the roles of each word.}{xp-activity-word}


Figure~\ref{fig:xp-activity-sentences} (Appendix~\ref{app:results-activity}) presents these same activities measured on four samples, with each plot showing activity for one sample. We can see that unit \texttt{U1}, specialized in positions, reacts when a position is present. Unit \texttt{U4} seems to react throughout the sentences, except to \textit{Positions}, with preferences for \textit{Objects} and \textit{Grammar}. Unit \texttt{U3} seems to remain empty until an \textit{Object} or a \textit{Color} appears. These examples allow us to visualize model activity word by word in a real situation, and seem to confirm the specializations seen above.


We also sought to determine whether the observed specialization is found across several seeds. As described in the experimental protocol, we trained 5 models on seeds 0 to 4, then aligned their memory units based on their predictive profiles. Figure~\ref{fig:xp-activity-aligned} presents the categories of the classes predicted by the linear readouts after aligning the units of the five models. For both situations, \textit{Positions} are mainly recovered within the same aligned unit \texttt{U1}. \textit{Objects}, \textit{Colors} and \textit{Relations} are more shared. In both situations, however, \texttt{U3} seems to capture \textit{Objects} and \textit{Colors}, \texttt{U2} \textit{Objects} and \textit{Relations}, and \texttt{U3} \textit{Colors} and \textit{Positions}. 

\xpfig{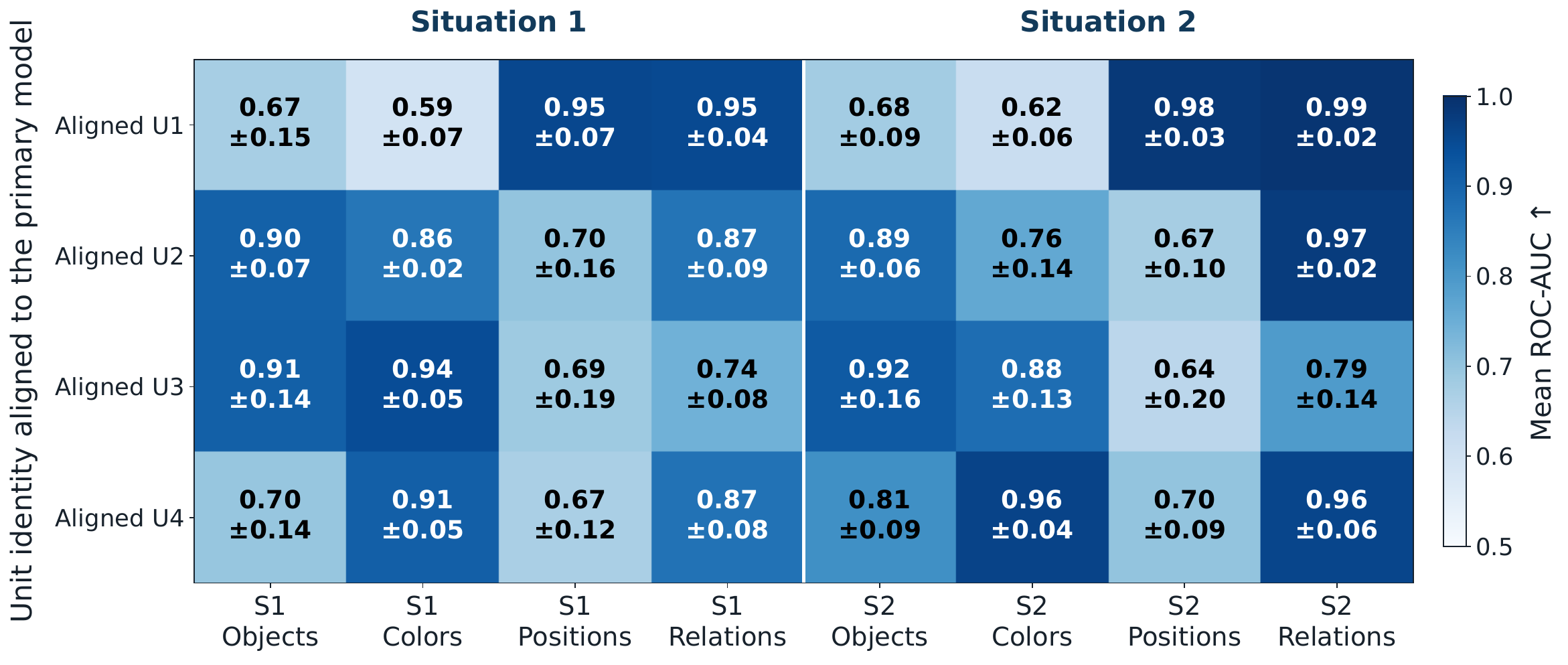}{ROC-AUC of the linear readouts for \textit{Objects}, \textit{Colors} and \textit{Positions} for both situations. Results are averaged over five seeds after aligning the units. ROC-AUC measures the readout's ability to distinguish the presence and absence of a class: a value of 0.5 corresponds to chance, and 1 to perfect separation.}{xp-activity-aligned}


To verify that all these units produce an output useful to the model, we zeroed their outputs one by one, without retraining the model. Figure~\ref{fig:xp-activity-ablation} shows that exact-match accuracy drops from 99.6\% to 6.0\%, 14.4\%, 26.1\% or 1.7\% depending on the unit removed. All four units therefore contribute to solving the task.

Finally, Figure~\ref{fig:xp-activity-similarity} (Appendix~\ref{app:results-activity}) compares the similarity of the final states and attention across units. Off-diagonal similarity values are low, between 0.02 and 0.13, while mean attention is well distributed across several units, with values of approximately 0.19 to 0.33 (a perfectly even distribution being 0.25 everywhere). We can therefore observe that the units produce different representations and that the decoder consults all these units. 

\xpfig{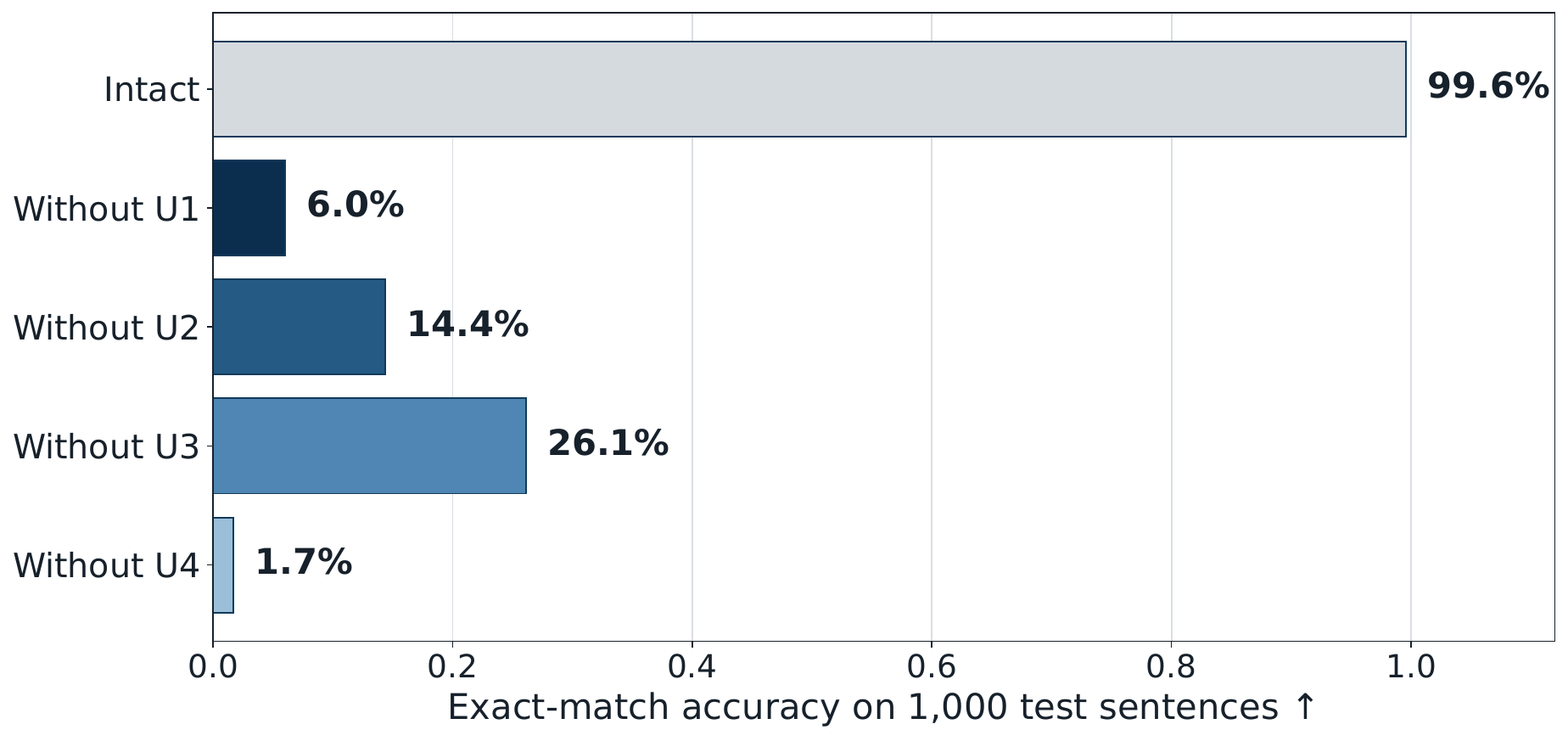}{Exact-match accuracy on the 1\,000 \texttt{cross\_situation} test examples when the output of one memory unit is zeroed throughout the sequence. Other parameters remain unchanged and the model is not retrained.}{xp-activity-ablation}



\subsection{Tiny Shakespeare}

\paragraph{Experimental protocol.}
To evaluate DMT for small-scale text generation, we started with the Tiny Shakespeare dataset. The model must learn to predict the next character from a vocabulary of 65 characters, with a context length of 64. The complete dataset contains just over 1.1M characters, or just over 17k samples once divided into windows. Samples are not shuffled and are split as follows: 85\% for training, 5\% for validation and 10\% for testing. The recurrent state is reset between samples.
We compare the three architectures with approximately 300k parameters and an arbitrary configuration (based on the literature for Mamba and Transformer). To find their training hyperparameters, we perform a grid search over six learning rates and four weight decays, with ten seeds ranging from 0 to 9. This represents 720 trained models. Training uses AdamW, one H100 GPU, a batch of 64, a learning rate with cosine decay and early stopping. We will first compare the learning curves of DMT, Mamba and Transformer, then their performance as a function of the number of epochs.

\paragraph{Results.}


For each architecture, we retained the learning rate/weight decay combination with the best mean validation accuracy. Figure~\ref{fig:xp-shakespeare-learning} presents the evolution of training loss, validation loss and validation accuracy for the ten selected seeds of each architecture. All three architectures learn the task, but Mamba and Transformer generally achieve a lower validation loss than DMT: 1.56 for Mamba and Transformer versus 1.61 for DMT, with mean test accuracy of 51.1\% for Mamba, 50.6\% for Transformer and 50.1\% for DMT. 
We can also note that, depending on the model, early stopping is triggered sooner or later. Thus, Mamba reaches convergence very quickly, in approximately 20 epochs, while DMT requires approximately 45 and Transformer is much slower, with 80 epochs.

\xpfig{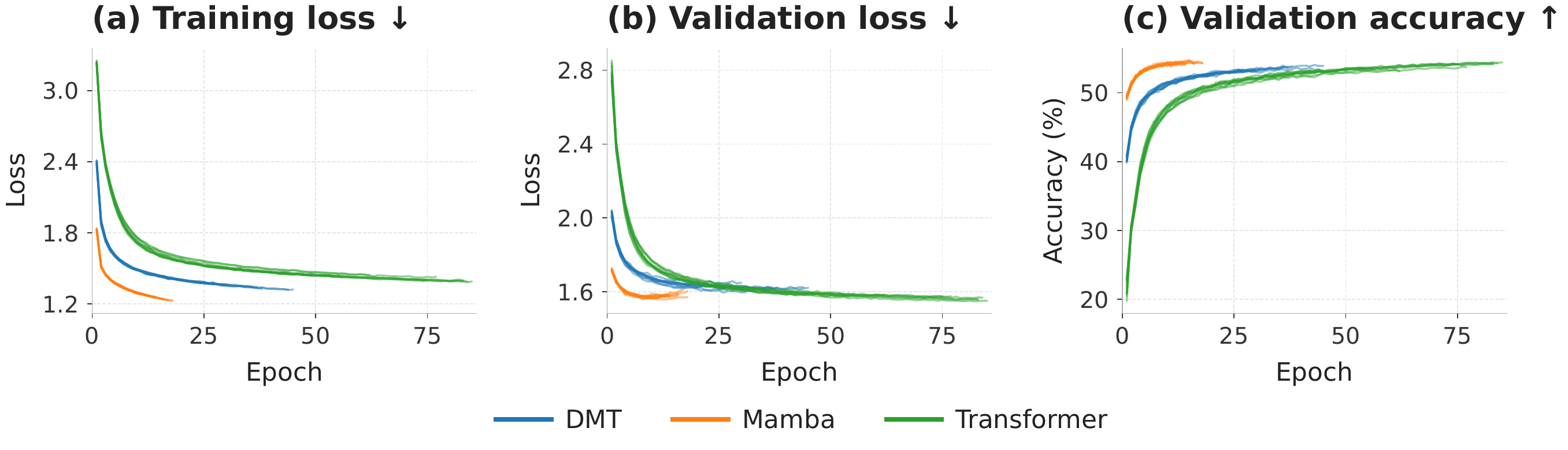}{Evolution of training loss, validation loss and validation accuracy on Tiny Shakespeare, for each of the ten seeds per architecture. The displayed learning rate/weight decay combination is the one with the best mean accuracy. Each curve stops at the run's early stopping point.}{xp-shakespeare-learning}


Figure~\ref{fig:xp-shakespeare-epochs} presents the test accuracy of the 720 models as a function of the number of epochs needed to reach convergence. The best results are not necessarily those obtained with the largest number of epochs. We can observe that, across all its models, Mamba achieves excellent results, all close to 50\% test accuracy, for training durations ranging from 8 to 60 epochs. DMT and Transformer seem to show two performance profiles (related to more or less suitable hyperparameter combinations). In particular, we observe that Transformer does not always succeed at the task, with some models well below 25\% accuracy, whereas DMT almost never falls below 40\%. For these two architectures, a minimum number of epochs seems necessary to achieve scores above 50\%, although DMT seems to converge much faster than Transformer overall.

\xpfig{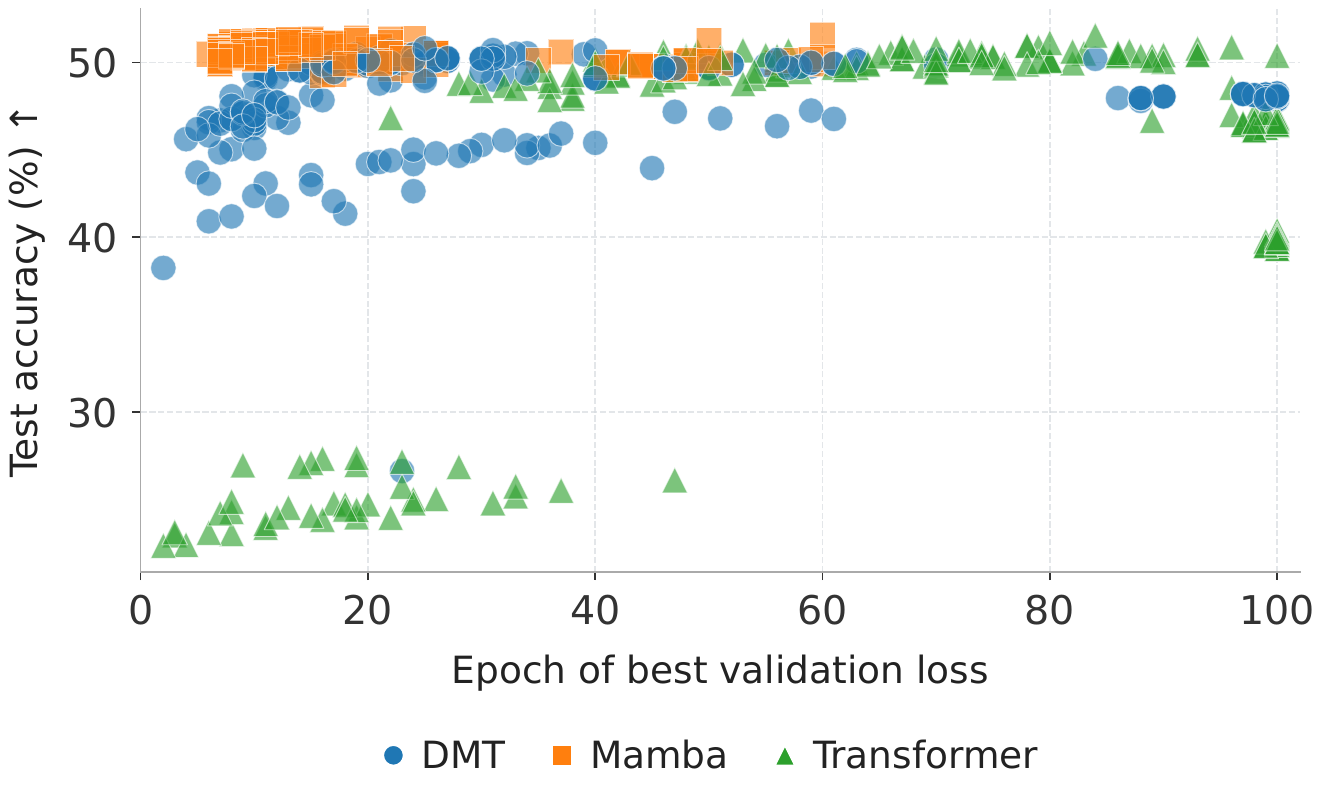}{Next-character prediction accuracy on the test set as a function of the number of epochs. Each point corresponds to one of the 720 trained models.}{xp-shakespeare-epochs}


\subsection{TinyStories}

\paragraph{Experimental protocol.}
We then evaluated the three architectures on TinyStories, with approximately 10M parameters and next-token prediction. We use a byte-level BPE tokenizer, which we trained ourselves with a vocabulary of 2\,048 tokens, and windows of 256 tokens on 1\% of the training set. The dataset contains a total of nearly 2.1M samples of 256 tokens; the first 98\% of samples form the training set, the next 1\% the validation set and the remaining 1\% the test set. The latter two therefore each represent nearly 21,000 samples. 
Training uses AdamW, four H100 GPUs, a batch of 256, a learning rate with cosine decay, cross-entropy and early stopping. Validation is performed every 500 samples. For each model, test perplexity is computed by taking the exponential of its test loss; we then average perplexities across seeds.
Unlike the CogScale experiment, here we separated the hyperparameter search and the architectural configuration search into several stages. We start by selecting an arbitrary 10M-parameter configuration for each architecture. We then perform a grid search over 24 learning rate/weight decay combinations on five seeds (0 to 4) per architecture, or 360 models. With the best combination found, and still using the same configuration, we then test 16 combinations of warmup and minimum learning rate, or 240 models.
The next step is to find the best configuration for each model. For this purpose, once these training hyperparameters are fixed, we test 60 configurations per architecture, all between 9 and 11M parameters. For DMT, the goal of this search is to discover which allocation of parameters between memory and attention is preferable for text generation.
The 60 configurations per architecture are first trained for five epochs on seed 0. We retain the best 15 per architecture, which we train on seeds 1 and 2, then five finalists, which we train until early stopping on seeds 0 to 4. This gives a total of 345 runs. Models were selected based on validation loss. To illustrate our comparisons and evaluate our models, we will use either validation loss or test perplexity.
The configurations explored, their parameter counts and their selection at the different stages are detailed in Appendix~\ref{app:tiny-configurations}.

\paragraph{Results.}

Figure~\ref{fig:xp-tiny-hp} (Appendix~\ref{app:results-tinystories}) presents the mean test perplexity obtained for each learning rate/weight decay combination during the first grid search. We can observe that the learning rate influences performance more than weight decay. For DMT, a high learning rate ($\ge 0.03$) leads to strong instability, with a high mean perplexity and standard deviation. A low learning rate ($\le 3e^{-4}$) drastically reduces performance. Transformer seems to be affected only by a low learning rate, and Mamba shows the most stable profile, although it is also affected by a learning rate that is too high or too low. The learning rate/weight decay combinations selected based on mean validation loss are $0.01/0.01$ for DMT, $0.003/0.1$ for Mamba and $0.01/0.01$ for Transformer.


Figure~\ref{fig:xp-tiny-seeds} (Appendix~\ref{app:results-tinystories}) then presents the five seeds for the selected hyperparameters. We can observe that, regardless of the seed tested, the same architecture achieves relatively similar scores. When a seed is difficult for one architecture, it seems to be difficult for the others as well (for example, seed 3), and conversely (for example, seeds 1 and 2). We will continue training 5 seeds for our tests, but this will justify the staged configuration search performed below to find the best among 60.


We then searched for the warmup and the lowest learning rate reachable during training (represented by the hyperparameter \texttt{min\_lr\_ratio}, a factor applied to the initial learning rate) suitable for each architecture. 
Figure~\ref{fig:xp-tiny-warmup} (Appendix~\ref{app:results-tinystories}) shows that DMT and Transformer become particularly unstable without warmup, with many runs diverging, whereas Mamba remains stable across all tested values, including without warmup. Although it can slightly influence the final score, the minimum value reachable by the learning rate does not seem particularly important. The warmup/\texttt{min\_lr\_ratio} pairs selected on validation are $0.30/0.01$ for DMT, $0/0.003$ for Mamba and $0.03/0.01$ for Transformer.


With the training hyperparameters selected, it is now possible to search for the best configuration for each architecture. As seen previously, a seed that is poor for one architecture seems to be poor for the others as well. Thus, to reduce the number of models to train and allow a configuration search with a controlled budget, we chose to progressively filter configurations by first limiting training to 5 epochs on seed 0, then on seeds 1 and 2, before performing complete training runs on all 5 seeds.  
Figure~\ref{fig:xp-tiny-stage1} (Appendix~\ref{app:results-tinystories}) presents the best 15 configurations among 60 for each architecture after five epochs on seed 0. This first stage reduces the number of configurations to test at the next stage to 15.


As with CogScale, since we have results for dozens of configurations per architecture, we sought to determine which architectural parameters were associated with better performance. Figure~\ref{fig:xp-tiny-features} presents Spearman correlations between these parameters and the negative of the best validation loss. We take the negative of the loss to obtain positive correlations when an architectural parameter reduces the loss.
We observe that DMT seems to prefer profiles with a large number of small-dimensional memory units; the number of decoder layers and its dimension are not necessarily correlated with better performance. Thus, DMT seems to favor encoder width over decoder depth. Mamba and Transformer have a more conventional profile, seeming to prefer the depth of their network over its width.

\xpfig{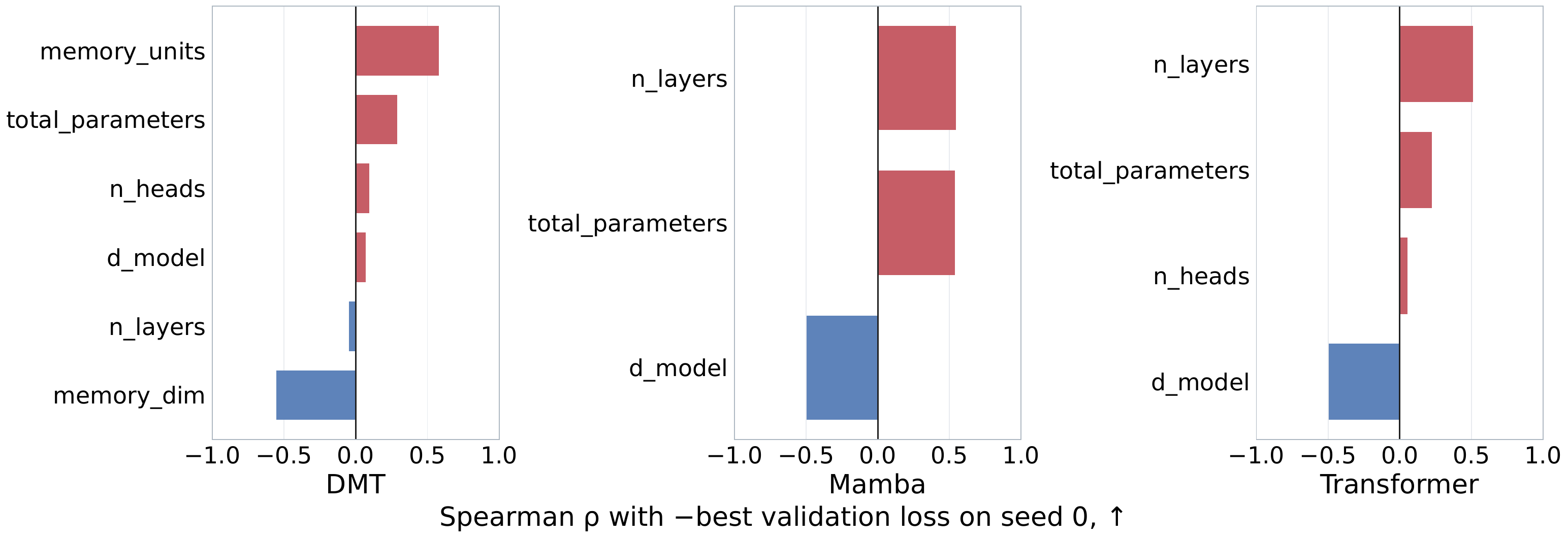}{Spearman correlation between architectural parameters and the negative of the best validation loss, after five epochs on seed 0. A positive correlation indicates an association with better performance.}{xp-tiny-features}


Figure~\ref{fig:xp-tiny-stage2} (Appendix~\ref{app:results-tinystories}) presents the mean validation loss of the 15 configurations over seeds 0, 1 and 2. We can observe the standard deviation across seeds here, which was not visible in the first ranking. Here too, training is limited to 5 epochs. This stage allows us to identify the 5 final configurations for each architecture.


Finally, Figure~\ref{fig:xp-tiny-stage3-val} (Appendix~\ref{app:results-tinystories}) presents the ranking of the five finalists after training on five seeds until early stopping. We select DMT \texttt{c21}, Mamba \texttt{c58} and Transformer \texttt{c45} based on their best mean validation loss. The differences obtained between these configurations remain small, not exceeding 0.01 loss between the best and worst for each architecture. Figure~\ref{fig:xp-tiny-stage3-test} (Appendix~\ref{app:results-tinystories}) presents the test perplexities of these same configurations, retaining the validation ranking order. We can note that the order is not always preserved between validation and test. Indeed, Transformer configurations \texttt{c54}, \texttt{c14} and \texttt{c35} switch order on the test set.


Finally, Figure~\ref{fig:xp-tiny-winners} compares the best configuration obtained for each architecture, with the validation results of each seed and their mean test perplexity. DMT \texttt{c21} achieves a mean perplexity of 4.29, compared with 3.60 for Mamba \texttt{c58} and 4.02 for Transformer \texttt{c45}. Their best mean validation losses are 1.47, 1.29 and 1.40, respectively. Despite the configuration search, DMT therefore remains behind the other two architectures on TinyStories, and Mamba achieves by far the best results.

\xpfig{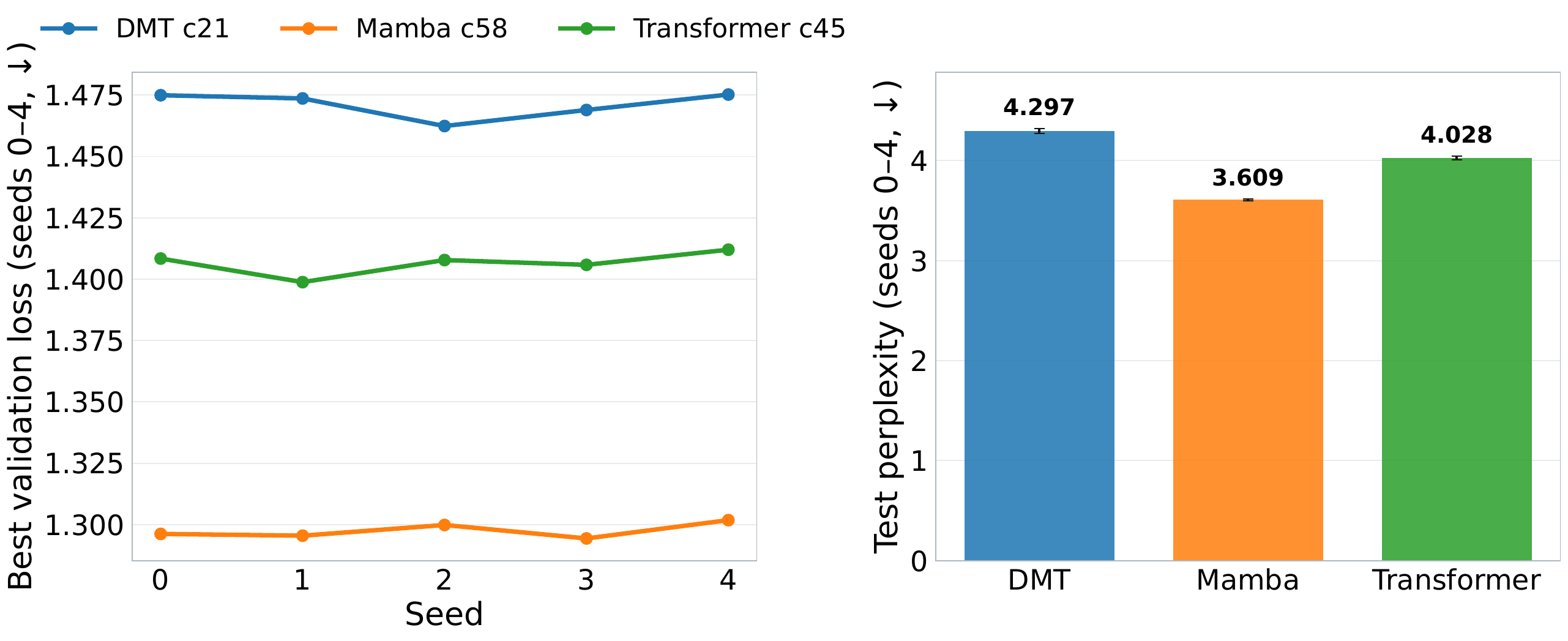}{Comparison of the best configuration obtained for each architecture. Left: the best validation loss for each seed from 0 to 4. Right: mean test perplexity and its mean standard deviation over the 5 seeds.}{xp-tiny-winners}


\subsection{OpenWebText}
\label{sec:owt}

\paragraph{Experimental protocol.}
To study scaling to a larger size, we trained the architectures on OpenWebText with a target budget of approximately 125M parameters. As explained in Section~\ref{sec:method}, our memory units in DMT approximate the actual state value by interpolation between the previous state and the candidate state (through the leak rate). As described by Equation~\eqref{eq:recurrence}, this interpolation is less precise than the solution used by Mamba, which aims to compute the actual state value if time were continuous, rather than an approximation. In this experiment, in addition to DMT, Mamba and Transformer, we wanted to evaluate another hybrid solution inspired by DMT, where we apply attention to memories implemented by Mamba SSMs. Overall, the resulting architecture is very similar to DMT, with the main difference being how the state update is computed. In the figures and text below, we will call this model Transba. 

The models use the GPT-2 tokenizer and its vocabulary of 50\,257 tokens, a context length of 1\,024 and an effective batch of 512 sequences, or 524,288 tokens per batch. The dataset is divided into 1\,000 shards, including 950 for training, one for validation and 49 for testing. DMT has 127.12M parameters, Mamba 129.12M, Transformer 124.34M and Transba 153.94M. Transba therefore exceeds the budget of the three main architectures. While Mamba, Transformer and Transba reach 950 shards, DMT training was stopped after 899 shards and was not resumed.\footnote{The reason is as follows: the run reached its maximum 20-hour allocation, and we did not have the resources to resume it. However, DMT had not finished learning. That said, the gain over the last 51 shards would certainly not have exceeded 0.02 validation loss points.} 
DMT and Transba use a maximum learning rate of $10^{-3}$, compared with $6\times10^{-4}$ for Mamba and Transformer.
Although 125M is still a small model compared with current LLMs, such scaling nevertheless requires a non-negligible amount of computation. Indeed, training was performed on 32 H100 GPUs. 
We therefore have only one run per architecture here, performed over a single epoch. 
For each architecture studied, we present its learning curve, final validation loss, final validation perplexity and next-token prediction accuracy on the 49 test shards. We also evaluate these architectures zero-shot on the WikiText-2, LAMBADA OpenAI, PIQA, ARC-Easy, HellaSwag, OpenBookQA and SciQ language benchmarks.
For multiple-choice tasks, we measure accuracy by choosing the answer with the highest total log-likelihood, then the one obtained after normalizing by answer length.


\paragraph{Results.}
Figure~\ref{fig:xp-owt-training} presents the results of the four runs. As with the TinyStories results, Mamba achieves by far the best validation loss, at 3.00, a validation perplexity of 20.15 and a test accuracy of 42.37\%. DMT ranks last, slightly behind Transformer, with a validation loss of 3.52, a perplexity of 33.93 and an accuracy of 36.54\%. The most interesting score comes from Transba, which very slightly beats Transformer, with a validation loss of 3.35 versus 3.38, a perplexity of 28.64 versus 29.27 and an accuracy of 38.89\% versus 38.63\%. That said, since Transba has 20\% more parameters than its competitors, this gain is too small to place it above Transformer. Thus, in this comparison, the DMT architecture and its Transba variant remain behind the two reference architectures, Mamba and Transformer.

\xpfig{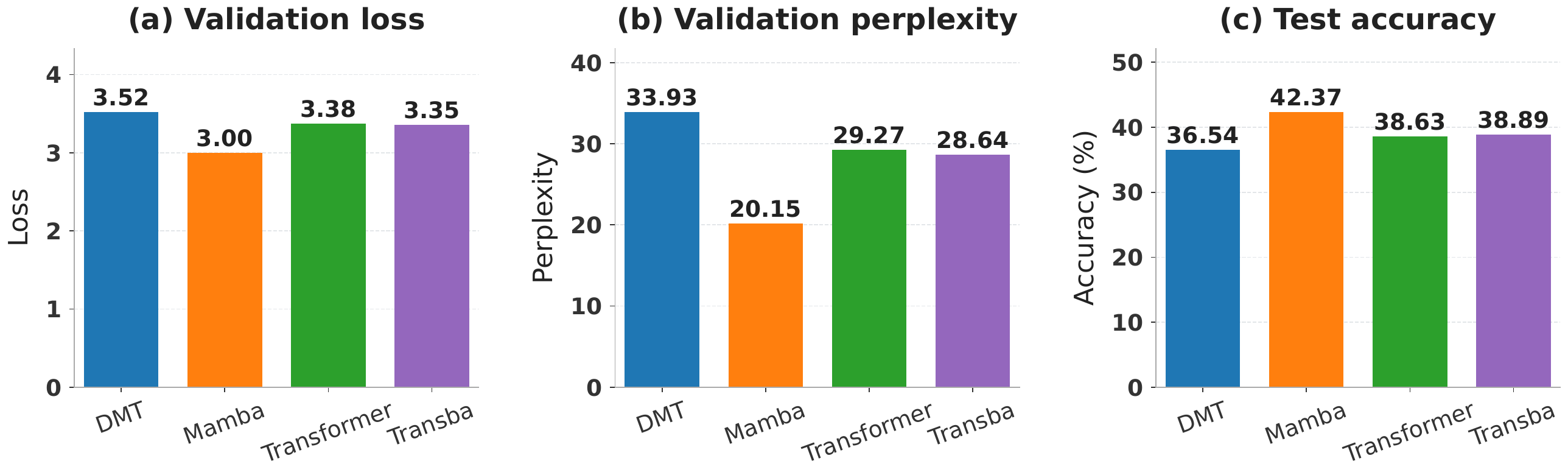}{Results on OpenWebText with one run per architecture: (a) validation loss, (b) validation perplexity and (c) next-token prediction accuracy on the 49 test shards.}{xp-owt-training}


Figure~\ref{fig:xp-owt-curves} (Appendix~\ref{app:results-owt}) presents the evolution of validation loss during training, with a zoom on the end of training. 
We can observe that Mamba's advantage is present from the start of training. Its curve is very smooth and drops faster and lower than the other models. DMT and Transba seem to follow the same slope at first, then diverge after 1.5B tokens, with Transba taking a better slope. We can note that DMT's curve seems slightly less smooth than Transba's, showing very slightly less stable training. Finally, we can note that Transformer training seems much more unstable than that of the other models. Its early training does not drop like the other architectures, accumulating a delay that it eventually catches up with by crossing DMT's curve around 3B tokens and progressively approaching Transba. If training had been performed on more data, Transformer would certainly have ended up surpassing Transba. However, it should be noted that the instability observed for Transformer (and, by extension, DMT) may be related to a poor selection of training hyperparameters for this dataset.


\xpfig{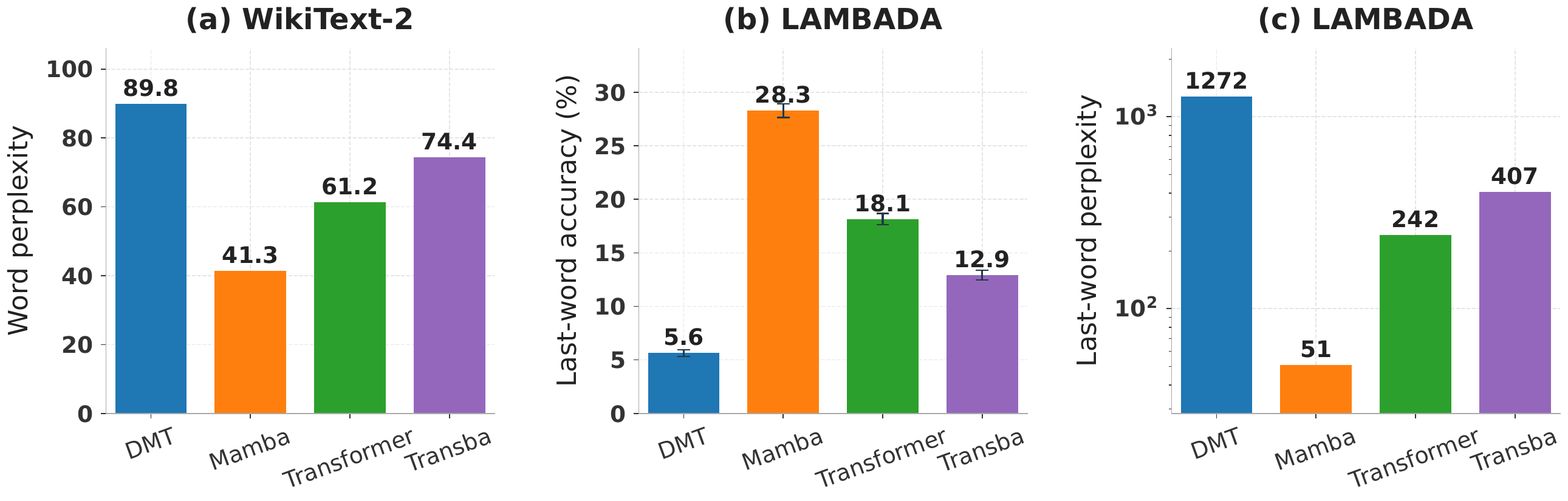}{Zero-shot evaluation of the models obtained after training on OpenWebText on the WikiText-2 and LAMBADA OpenAI language benchmarks. }{xp-owt-language}

Figure~\ref{fig:xp-owt-language} presents the results obtained on the WikiText-2~\citep{merity2017pointer} language modeling benchmark, which evaluates the prediction of text from Wikipedia articles, and LAMBADA~\citep{paperno2016lambada}, which asks for the last word of a narrative passage to be predicted. For WikiText-2, we measure perplexity: the lower it is, the better the model predicts the text. For LAMBADA, we measure perplexity and the accuracy of the word to predict. 
We observe that Mamba retains first place. In particular, we note that despite worse scores than Transba during training, Transformer takes second place on these benchmarks, well ahead of Transba and well behind Mamba. It therefore seems to have learned to generalize much better to data whose distribution is not part of training. Finally, as could be expected from the training results, DMT ranks last on both tasks. It seems less far behind on WikiText-2 than on LAMBADA. This may be explained by some of the texts in our training dataset (OpenWebText) coming from Wikipedia. Thus, DMT seems to have generalized much less than its competitors.


Finally, Figure~\ref{fig:xp-owt-choice} compares results on five multiple-choice tasks: PIQA, which tests physical knowledge \citep{bisk2020piqa}; HellaSwag, which asks for the likely continuation of a situation \citep{zellers2019hellaswag}; ARC-Easy~\citep{clark2018arc} and OpenBookQA~\citep{mihaylov2018openbookqa}, which ask scientific questions; and SciQ, which asks for useful information to be retrieved from a passage to answer a scientific question as well \citep{welbl2017sciq}. The displayed results are the model's accuracy per benchmark over all questions asked.
Here too, Mamba consistently achieves the best score on all benchmarks. That said, it does not necessarily outperform its competitors by a large margin. Surprisingly, DMT achieves relatively good scores, regularly in second place behind Mamba, as is the case for PIQA (tied with Transba), ARC-Easy and OpenBookQA. In contrast, when information must be retrieved from a passage, as with SciQ, it ranks last. We also note, surprisingly, that Transformer does not particularly excel on these tasks, regularly ranking last. Transba has a fairly varied profile, between DMT and Transformer. Thus, although DMT's training on OpenWebText did not lead to generalization as advanced as the other architectures, it nevertheless produced enough factual knowledge to achieve respectable scores on PIQA, ARC-Easy and OpenBookQA. 

\xpfig{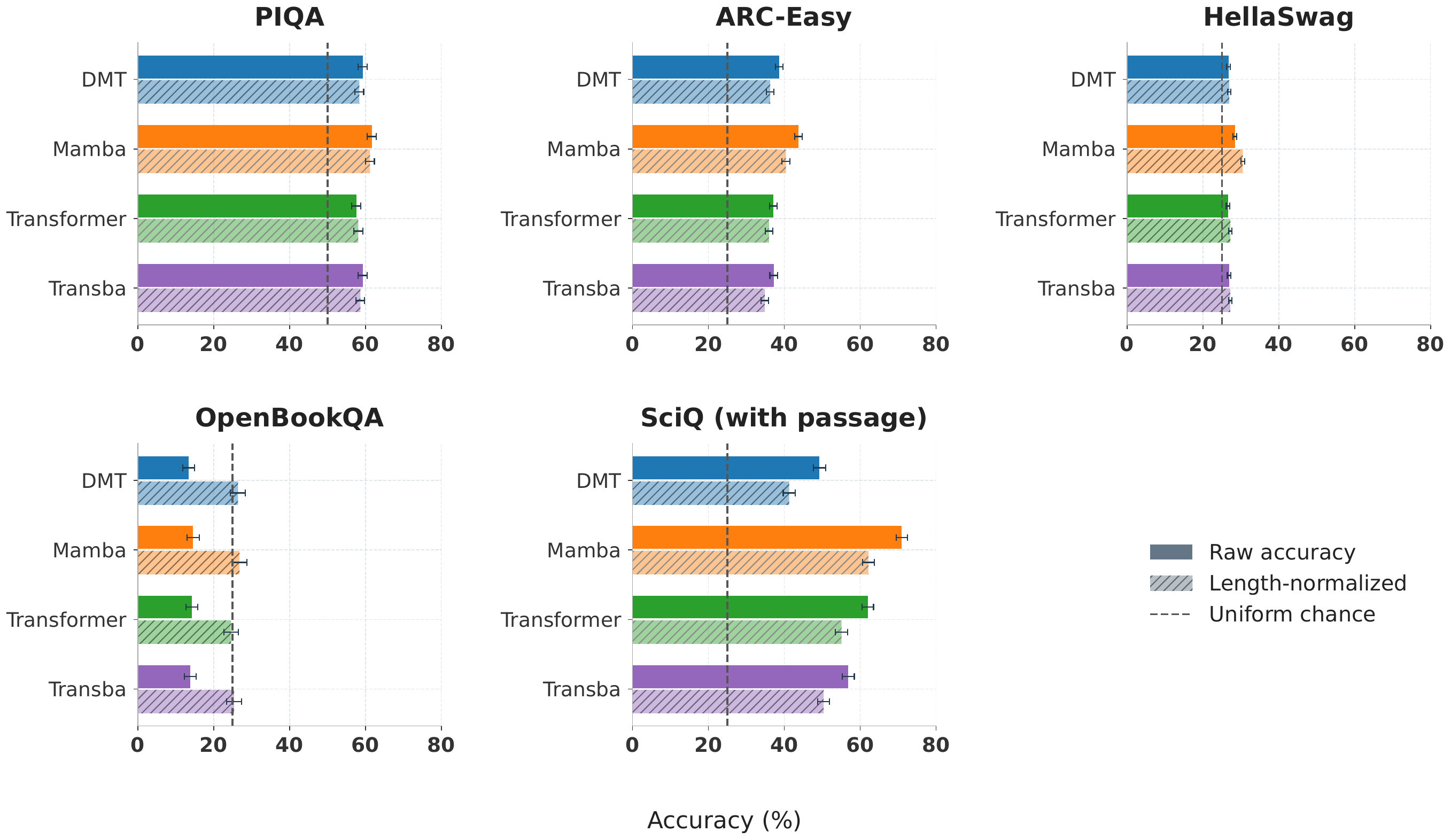}{Zero-shot evaluation of the models obtained after training on OpenWebText on the PIQA, ARC-Easy, HellaSwag, OpenBookQA and SciQ language benchmarks.}{xp-owt-choice}

\subsection{Computational cost and memory}
\label{sec:compute-cost}

\paragraph{Experimental protocol.}
We then measured the architectures' memory and time costs during training, initial context processing and autoregressive inference. We measure GPU time, peak allocated hardware memory and the size of the memory medium retained between two steps. We use DMT c21, Mamba c58 and Transformer c45, which we previously selected on TinyStories validation. 
We add DMT c05, the fastest of the 60 configurations we tested, and DMT c03, the one that maximizes the number of units and their dimension. In this way, we study DMT with different configurations. 
For this test, we use one H100 80 GB GPU, random weights and tokens, in FP32/complex64. We progressively test seven lengths, from 128 to 8\,192 tokens. Training includes the forward pass, cross-entropy, backpropagation, clipping and AdamW, with a batch of 4. Initial context processing uses the entire sequence with a batch of 1. Generation is performed over 32 tokens one by one, reusing the recurrent state or KV-Cache, so the initial context cost is not included in generation time.
To ensure that the GPU is properly initialized, we apply 5 warmup passes before taking our measurements over 20 repetitions. Each point is the median time obtained over these 20 repetitions.
We note that Transformer does not use FlashAttention.


\paragraph{Results.}
Figure~\ref{fig:xp-compute-time} shows that the cost of training and initial context processing depends on sequence length and model configuration. As expected, since Mamba and DMT are both SSMs with similar complexity, their curves are parallel. That said, DMT's cost, largely dominated by its $M^2$ attention, is generally much higher than Mamba's. For training, this appears once sequence lengths exceed $T \ge 256$. For initial context processing, it appears beyond $T \ge 1024$. DMT therefore seems to saturate a GPU's parallel capacity much faster than Mamba; this is particularly visible in the right plot, where Mamba seems to have a stable initial context processing time until sequence lengths exceed $T \ge 4096$. We can also observe the times obtained by Transformer in this figure, which, with its quadratic complexity in $T$, keep growing as $T$ increases, eventually exceeding DMT's cost around $T \ge 1024$. 

\xpfig{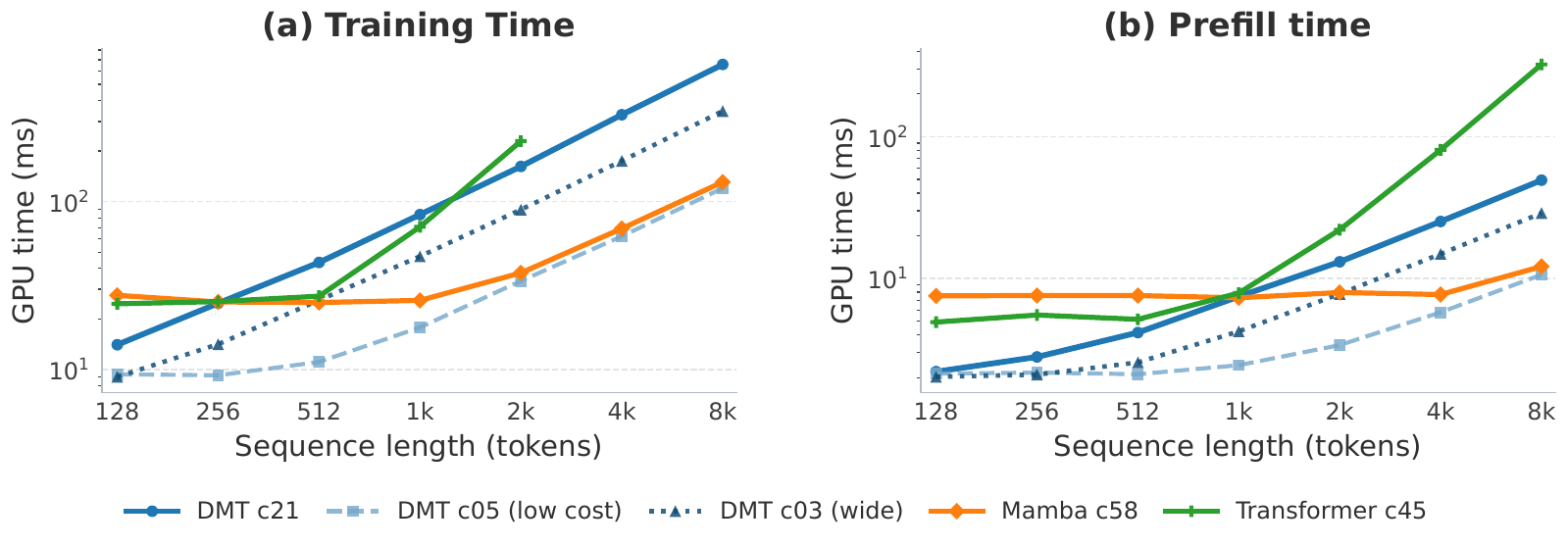}{Median GPU time for a training pass with a batch of 4 (left) and initial context processing with a batch of 1 (right). Transformer exceeds the available VRAM at the two largest lengths during training.}{xp-compute-time}


Figure~\ref{fig:xp-compute-memory} presents peak hardware memory allocated by PyTorch. This measure is mainly influenced by the number of values to retain in memory, which is closely related to the amount of computation performed. It is therefore not surprising that the curves seen here are very close to the previous GPU time curves. Here too, Mamba and DMT seem to follow parallel curves, with a strong advantage for Mamba. Similarly, Transformer's hardware memory cost exceeds DMT's for $T \ge 750$ during training and $T \ge 2048$ during initial context processing.  

\xpfig{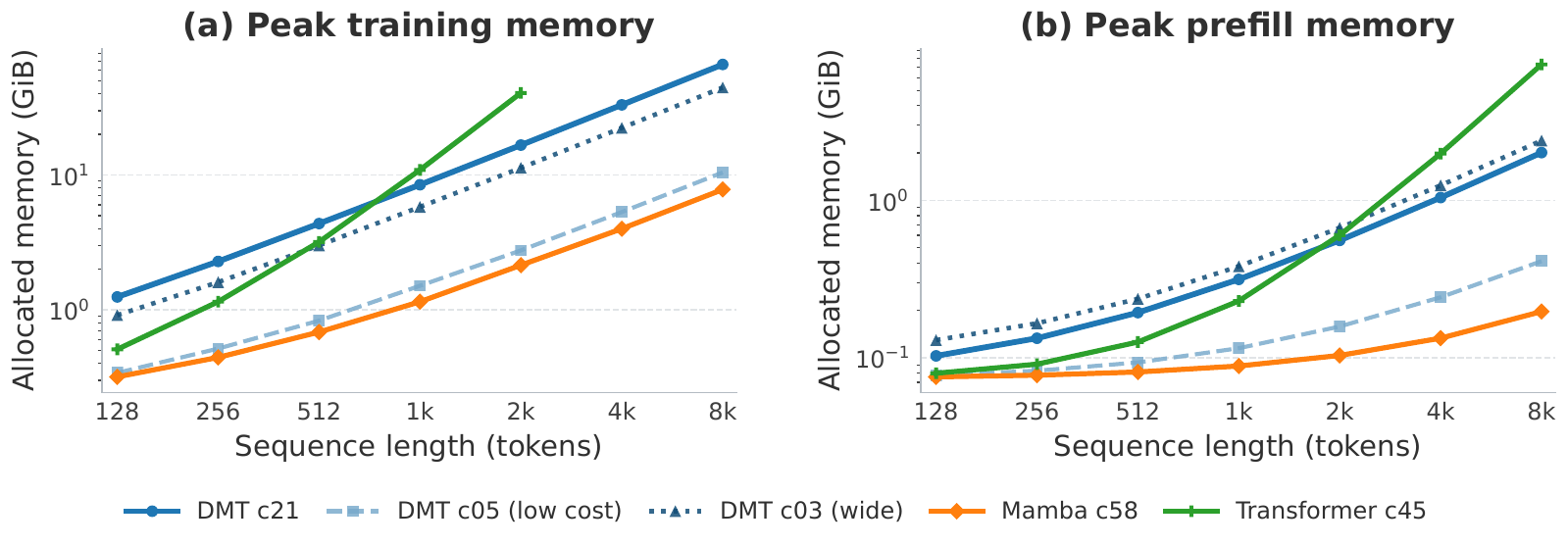}{Peak allocated hardware memory during the training update (left) and initial context processing (right).}{xp-compute-memory}


Finally, Figure~\ref{fig:xp-compute-decode} shows the time taken to generate a token and the size of the state used (recurrent state or KV-cache) as a function of sequence length. DMT clearly wins on both plots. First in latency, at approximately 1.6 ms per token for DMT c21, compared with 5.0 ms for Mamba and 3.9 ms for Transformer. And in its constant hardware memory cost of 68 KB for DMT c21, compared with approximately 1000KB, also constant, for Mamba and a growing, consistently higher cost for Transformer.   

\xpfig{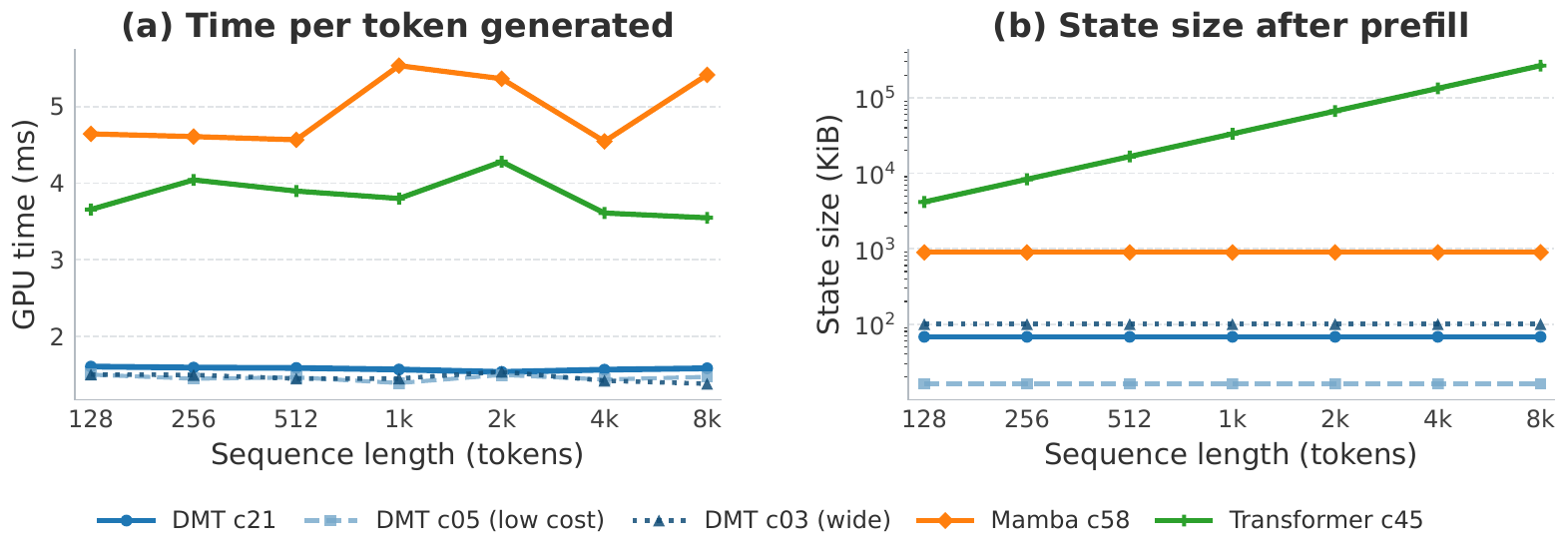}{Time to generate a token during autoregressive inference, with a batch of 4 (left), and size of the recurrent state or KV-cache after initial context processing (right), with a batch of 1.}{xp-compute-decode}


\subsection{Ablations}

\paragraph{Experimental protocol.}
To measure the role of the different DMT c21 modules, we remove the adaptive leak rate, temporal convolution and attention, separately and then simultaneously. Removing the adaptive leak rate amounts to using a fixed, learned leak rate. For attention, we retain the feedforward layers that follow the QKV attention layers. Thus, while the DMT c21 decoder initially had 4 layers composed of an attention layer followed by a feedforward layer, in this experiment it has only 4 feedforward layers. The removed weights are not reallocated elsewhere in the model. Thus, removing the adaptive leak rate removes 5,120 parameters, convolution 20,480 and attention 1,638,400 out of a total of 10,851,792 parameters.
The seven variants are trained on three seeds (0 to 2) for 5 epochs, then compared with the first five epochs of the corresponding complete runs.

\paragraph{Results.}

Figure~\ref{fig:xp-ablation-effects} shows that all variants perform worse than the corresponding complete variant. That said, not all modules seem to have the same value. Indeed, while the adaptive leak rate does not seem to greatly affect validation loss, which goes from 1.47 to 1.48, convolution seems to play a major role, with validation loss going from 1.47 to 1.59. Yet it represents only 0.2\% of the model's parameters. We now come to one of the most interesting results of this work: the contribution of attention in this model. While we can observe that removing this module significantly increases validation loss, from 1.47 to 1.52, we must also note that we removed 1,638,000 parameters from the model, or more than 16\% of it. Thus, proportionally, attention does not seem to play an important role in DMT. However, it is the layer that requires the most computation as the number of memory units $M$ grows, and therefore the most limiting one for scaling the architecture. 

\xpfig{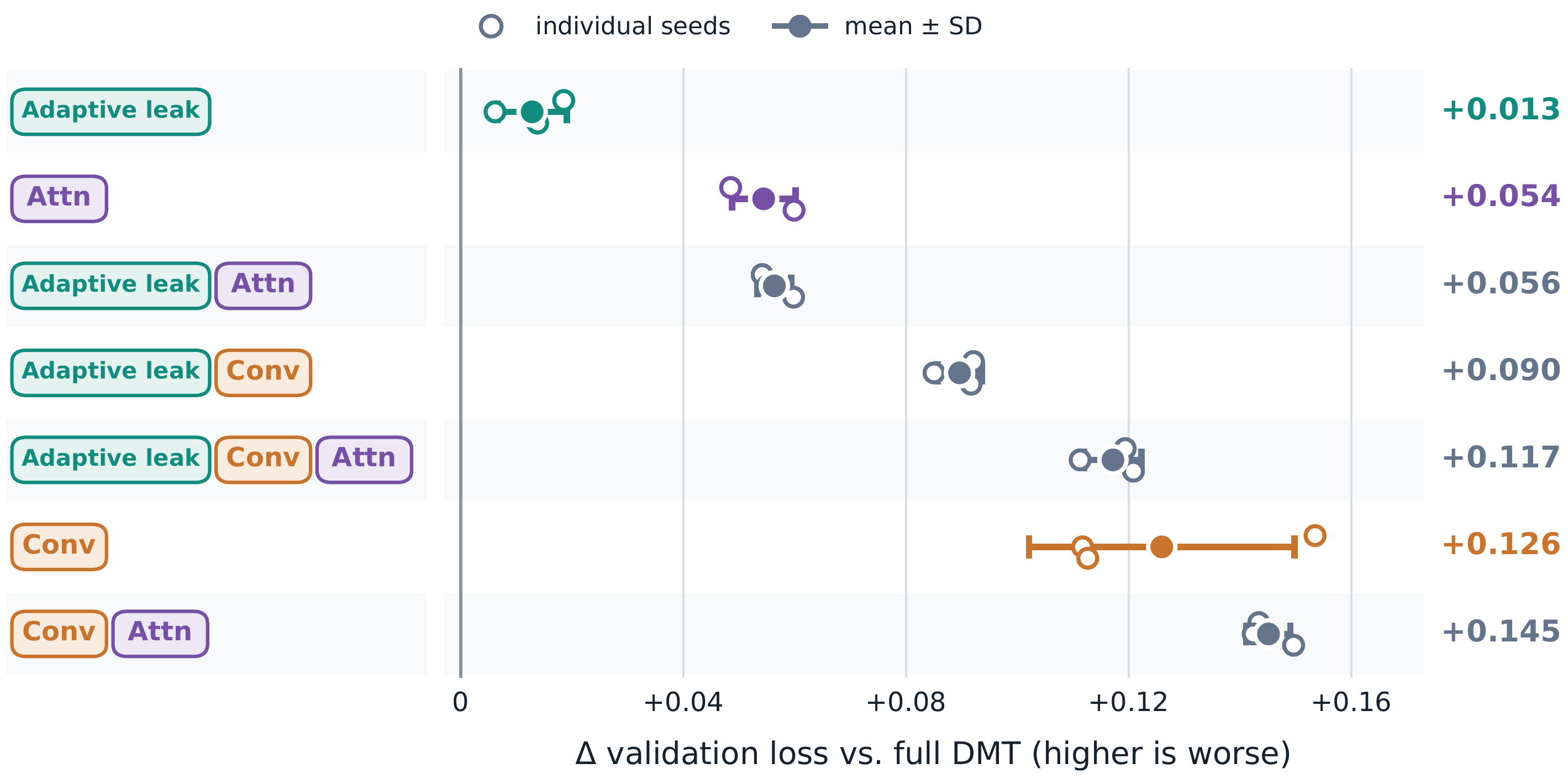}{Difference in validation loss relative to complete DMT after five epochs, paired by seed. Circles show the three seeds; solid points and bars indicate the mean and standard deviation.}{xp-ablation-effects}




Finally, Figure~\ref{fig:xp-ablation-time} shows the training times required for each variant. Unsurprisingly, the complete variant takes the longest. Those without adaptive leak rate and convolution are slightly faster, and those without attention are much faster, going from nearly 4 hours of training to less than 2 hours 30 minutes.  

\xpfig{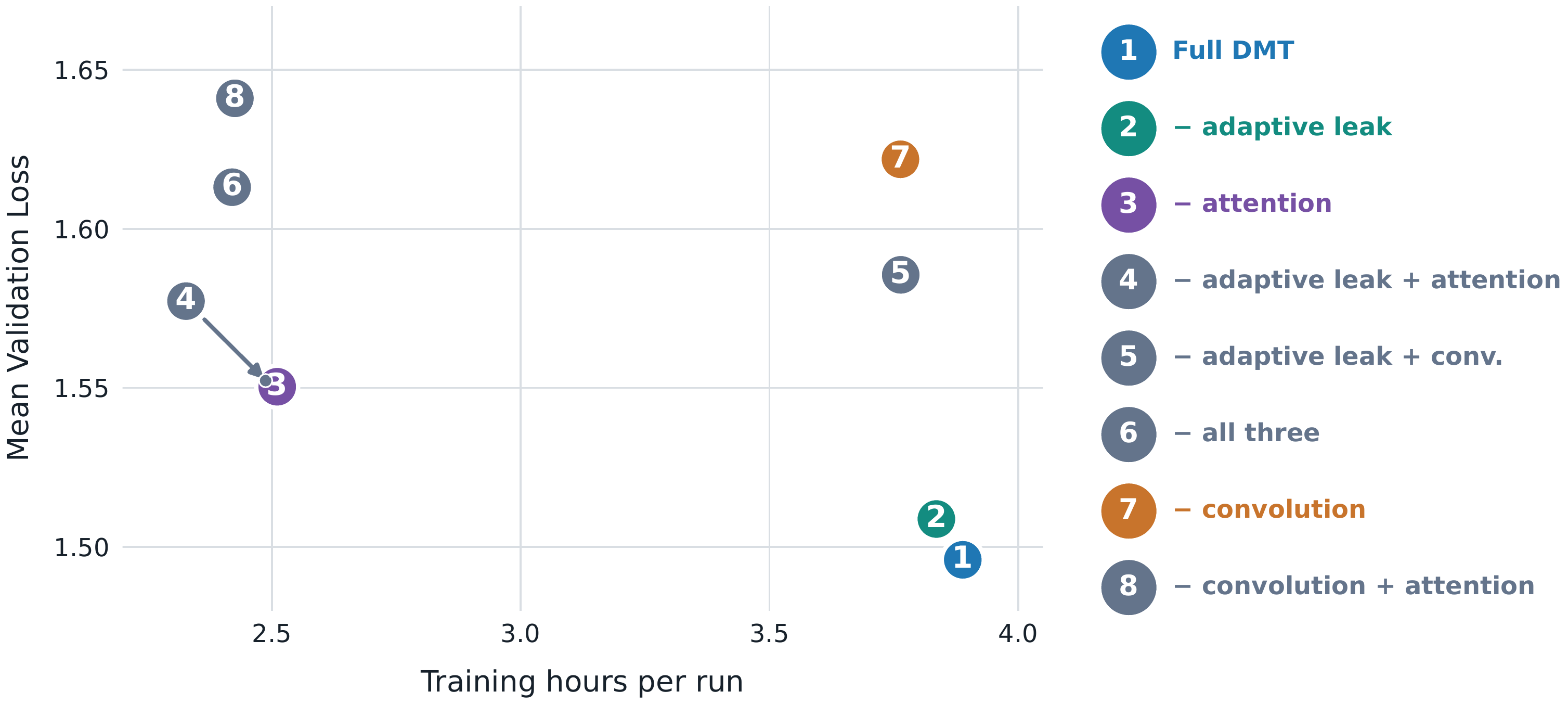}{Mean validation loss as a function of total time for five epochs for all tested variants and the original complete variant. Variants 3 and 4 are superposed. We moved variant 4 up-left its real position. The arrow indicates where it should really be.}{xp-ablation-time}

\section{Discussion}

\paragraph{What the experiments show.}
The experiments show that several recurrent memory units in parallel can encode a sequence and allow next-token prediction, while retaining a fixed-size state during inference. DMT is able to learn different cognitive tasks, shows activity that differs across its memory units and can extend learned rules beyond the training length. Our language modeling experiments show that this organization also works at different scales, from a few hundred thousand to approximately 125 million parameters. Nevertheless, these results do not place DMT ahead of the reference architectures: Mamba achieves by far the best language modeling results. That said, although slightly inferior, DMT competes closely with Transformer, while having fixed incremental complexity thanks to its fixed-size memory and logarithmic parallel depth in sequence length thanks to the associative scan.

\paragraph{State interpolation and discretization.}
Mamba and DMT, both SSMs, use different methods to update the state. Whereas
DMT uses the leak rate to interpolate between the previous state and a candidate state, Mamba constructs its update from the discretization of a continuous dynamical system. By definition, Mamba is therefore more precise than DMT in discretizing the system. That said, 
our results show that an interpolation-based update is sufficient for discrete problems, from cognitive tasks to text modeling. Furthermore, the tests performed with Transba on OWT do not provide a sufficient gain to establish that an interpolation-based update is inferior to the discretization used by Mamba.  

\paragraph{What role for the adaptive leak rate and attention?}
The ablation shows that the adaptive leak rate and attention can be questioned. While it is true that the adaptive leak rate does not seem necessary with 8 memory units, we think it may prove useful as the number of memory units increases. Indeed, a leak rate of 0 over several time steps can freeze a unit's state over a long period. We therefore think that, in addition to allowing dynamic selectivity, this mechanism may allow better information retention. That said, when the number of memory units is relatively small, it becomes very difficult to freeze units, since this reduces the state capacity available for subsequent incoming information.
Unexpectedly, attention also seems superfluous. Yet, for only 16\% of the studied configuration's weights, it accounts for nearly 40\% of the computation. Future experiments should therefore replace these attention layers with other mechanisms, less costly than $M^2$ attention. Although more suitable solutions must exist, the most straightforward would be to replace them with simple feedforward layers. The decoder would then be greatly simplified, consisting only of feedforward layers. Training and inference times would then be greatly reduced. 

\paragraph{Width and depth.}
In the configurations we explored, DMT seems better when increasing the number of memory units in parallel rather than the number of decoder layers. These results suggest that the diversity of our units' dynamics plays an important role. Furthermore, we can note that depth could be used differently. In the current architecture, all decoder layers read the encoder outputs. 
It may be possible to stack several blocks, each containing memory units, to perhaps obtain better temporal representations.

\paragraph{The importance of local context.}
Temporal convolution is the mechanism whose ablation reduces performance the most on TinyStories. It therefore seems to provide computations complementary to those of the recurrence. Indeed, it allows patterns to be extracted from successive tokens, which will then be provided to the different memory units. This operation is particularly relevant for language, where the information carried by a token often depends on its immediate context.
That said, it is not necessary for all tasks. 
When related elements are far apart or their neighbors are independent, this local processing is less useful. 

\paragraph{Computational cost and implementation.}
For computational cost, we must distinguish training, initial context processing and autoregressive generation. We show that DMT provides fast autoregressive generation compared with its competitors and a fixed hardware memory cost, regardless of context length. However, its attention, quadratic in the number of memory units $M$, makes it more costly than Mamba during training and initial context processing, both in time and hardware memory. 
That said, the implementation leaves room for optimization. Our measurements were performed using Triton to implement our CUDA kernel. As with Mamba, better time and memory performance could be observed with a CUDA kernel written in C++, but at the cost of more complicated installation and more specific hardware requirements.

\paragraph{Quality of the generated language.}
In agreement with the results on the language benchmarks evaluated after training on OWT, DMT's generations generally remain less interesting than Mamba's. Although DMT manages to retain context over a very large window, the resulting texts are less understandable and meaningful. Repetitions are more frequent. Examples in Appendix~\ref{app:owt-generations} illustrate these differences.

\paragraph{What the comparison with Mamba teaches us.}
Throughout this study, Mamba achieved very good results, very often ranking first. This shows how solid a foundation it provides for sequential processing. 
Its language results, training cost and autoregressive inference cost demonstrate the value of selective recurrent architectures for language modeling, particularly when compared with Transformer. That said, its difficulties on CogScale show that it is not always the best option, particularly in some conditions where the budget is very small.


\section{Conclusion}

We proposed Dynamical Memory Transformer (DMT), a hybrid architecture inspired by Reservoir Computing and Transformers that combines several memory units with an attention decoder. These units are adaptive dynamical systems that encode the sequence in parallel. By applying attention to their outputs rather than to past tokens, we obtain attention whose cost does not depend on context length. The associative scan allows state computations to be parallelized along the temporal dimension, while autoregressive inference retains a fixed-size state and a constant cost per new token.
The experiments show that this architecture can learn different cognitive tasks and model language at several scales. DMT achieves results competitive with those of Transformer, without surpassing it in text modeling. Mamba retains a clear advantage in most comparisons. Nevertheless, DMT shows an interesting profile by achieving the best results on several CogScale cognitive tasks.
Finally, the ablations invite us to reconsider some architectural choices. While temporal convolution plays an important role, the adaptive leak rate and attention provide a more limited gain. 
An interesting direction could therefore be to replace these mechanisms with others that are more suitable. These observations encourage further study of parallel memory units, seeking a simpler and less costly way to use their representations.

\clearpage
\bibliography{references}
\bibliographystyle{tmlr}

\clearpage

%% file: appendices.tex
\appendix

\section{Proof of the exact recurrence and scan}
\label{app:scan}

For completeness, let $p_i=(a_i,b_i)$ represent $f_i(h)=a_i\odot h+b_i$. According to Equation~\ref{eq:operator},
\begin{align}
  (p_1\otimes p_2)\otimes p_3
  &= (a_3a_2a_1,\;a_3(a_2b_1+b_2)+b_3),\\
  p_1\otimes(p_2\otimes p_3)
  &= (a_3a_2a_1,\;(a_3a_2)b_1+(a_3b_2+b_3)).
\end{align}
The results are equal by associativity and distributivity of elementwise complex arithmetic. The identity element is $(\mathbf{1},\mathbf{0})$. For $h_0=0$, the second component of the inclusive prefix $p_1\otimes\cdots\otimes p_t$ equals $h_t$. If $h_0$ is nonzero, we can prepend $(\mathbf{1},h_0)$ or add the accumulated transition multiplied by $h_0$. Non-commutativity is visible, for example, with $a_1=2,b_1=1,a_2=3,b_2=0$: the two orders give constant terms of three and one.

\section{Architectural and training details}
\label{app:architecture}

\begin{table}[H]
\centering
\small
\begin{tabular}{lccc}
  \toprule
  Setting & \DMT{} & Mamba & Transformer \\
  \midrule
  Tiny Shakespeare parameters & 306\,053 & 298\,560 & 304\,000 \\
  Tiny Shakespeare layers & 4 decoder blocks & 9 Mamba blocks & 6 decoder blocks \\
  Tiny Shakespeare context / vocabulary & 64 / 65 & 64 / 65 & 64 / 65 \\
  TinyStories parameters & 10\,851\,792 & 10\,901\,760 & 10\,871\,120 \\
  TinyStories layers & 4 decoder blocks & 21 Mamba blocks & 20 decoder blocks \\
  TinyStories context / vocabulary & 256 / 2\,048 & 256 / 2\,048 & 256 / 2\,048 \\
  \bottomrule
\end{tabular}
\caption{Configurations selected for language modeling.}
\end{table}

The selected TinyStories DMT c21 configuration uses $M=16$, $N=64$, $D=320$, four decoder layers and ten attention heads. The implementation applies RMS normalization to memory inputs, initializes leak rate biases using target rates distributed log-uniformly between 0.01 and 1, initially places $\lambda_m$ on the unit circle and applies SiLU after convolution, state reading and the pre-output projection. During autoregressive inference, it retains the last recurrent state and the $K-1$ previous convolution inputs of each memory.


\section{Supplementary results}
\label{app:complementary-results}

\subsection{CogScale}
\label{app:results-cogscale}

\xpfig{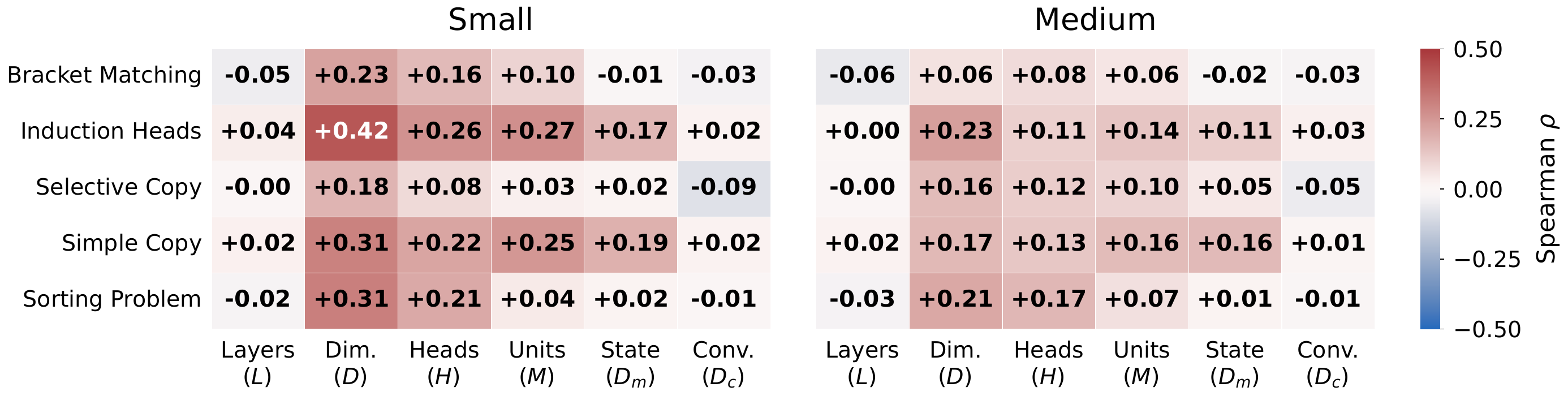}{Spearman correlation between DMT's different architectural parameters and performance obtained on CogScale, across all budgets, hyperparameters and configurations.}{xp-cogscale-hp}

\subsection{Memory activity on Cross Situation}
\label{app:results-activity}

\xpfig[width=1.2\linewidth]{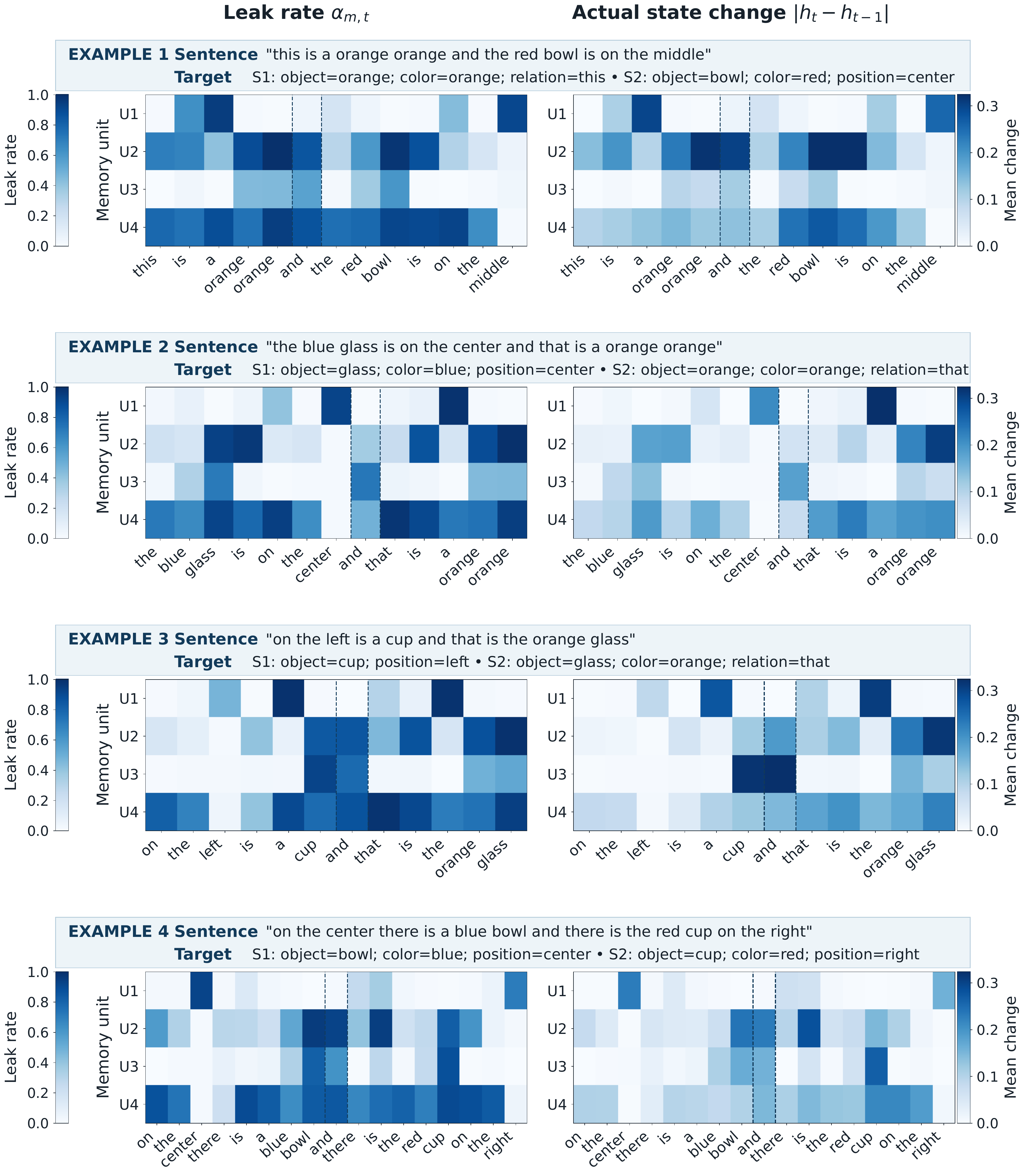}{Evolution of leak rate and state variation over four correctly solved \texttt{cross\_situation} sentences. Each plot presents one sentence and the two associated activities. Dashed lines delimit the connector between the two situations.}{xp-activity-sentences}

\xpfig[width=0.8\linewidth]{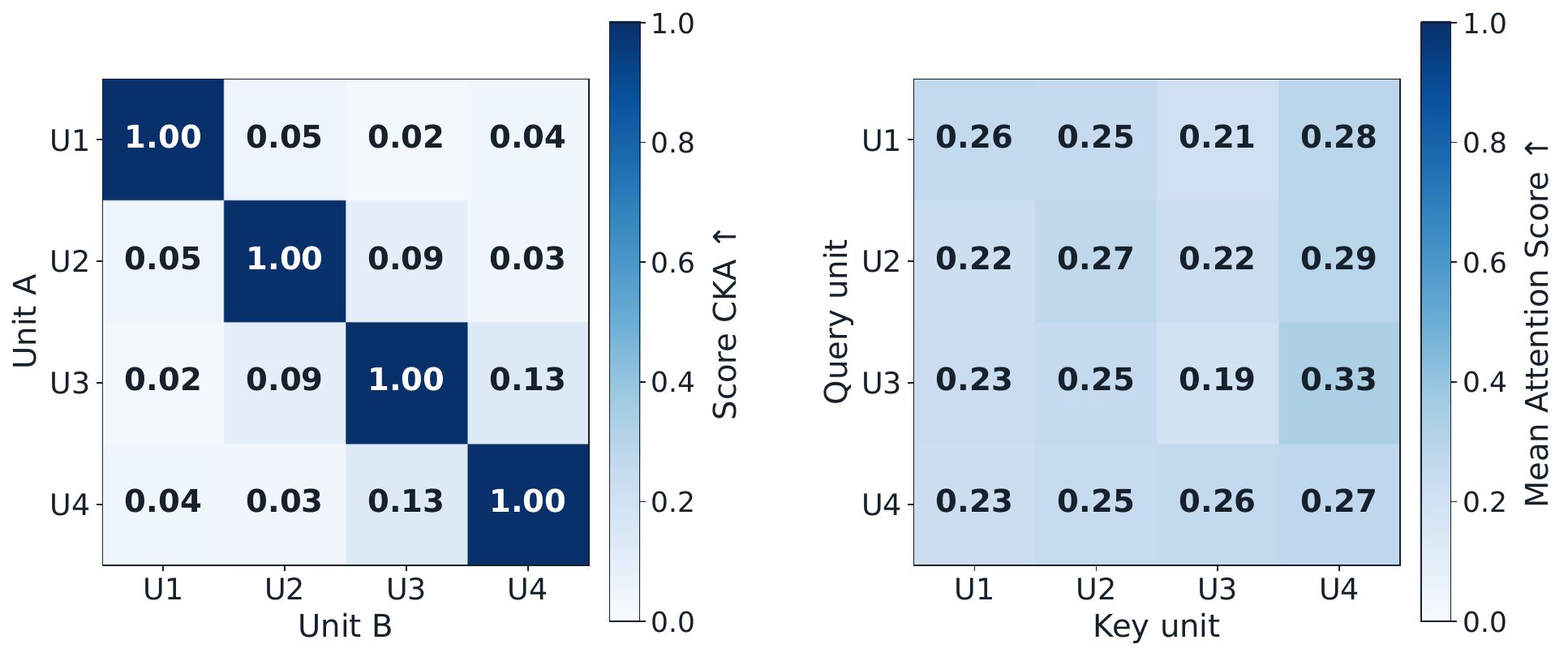}{Similarity of the final states of the memory units, measured by linear CKA on the test set (left), and mean decoder attention across units (right). CKA measures similarity between two sets of representations: a value close to 0 indicates low similarity, while a value close to 1 indicates high similarity.}{xp-activity-similarity}

\subsection{TinyStories}
\label{app:results-tinystories}

\xpfig{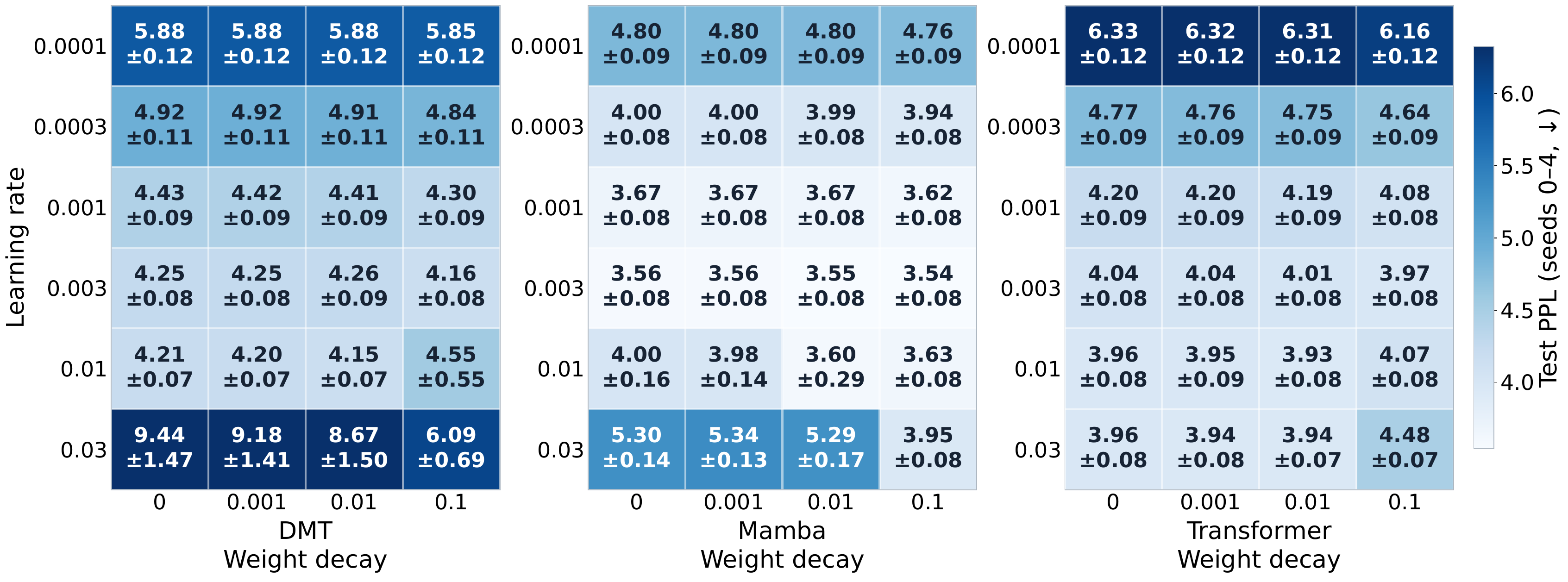}{Mean test perplexity and standard deviation over five seeds for each learning rate/weight decay combination and each architecture.}{xp-tiny-hp}

\xpfig[width=0.8\linewidth]{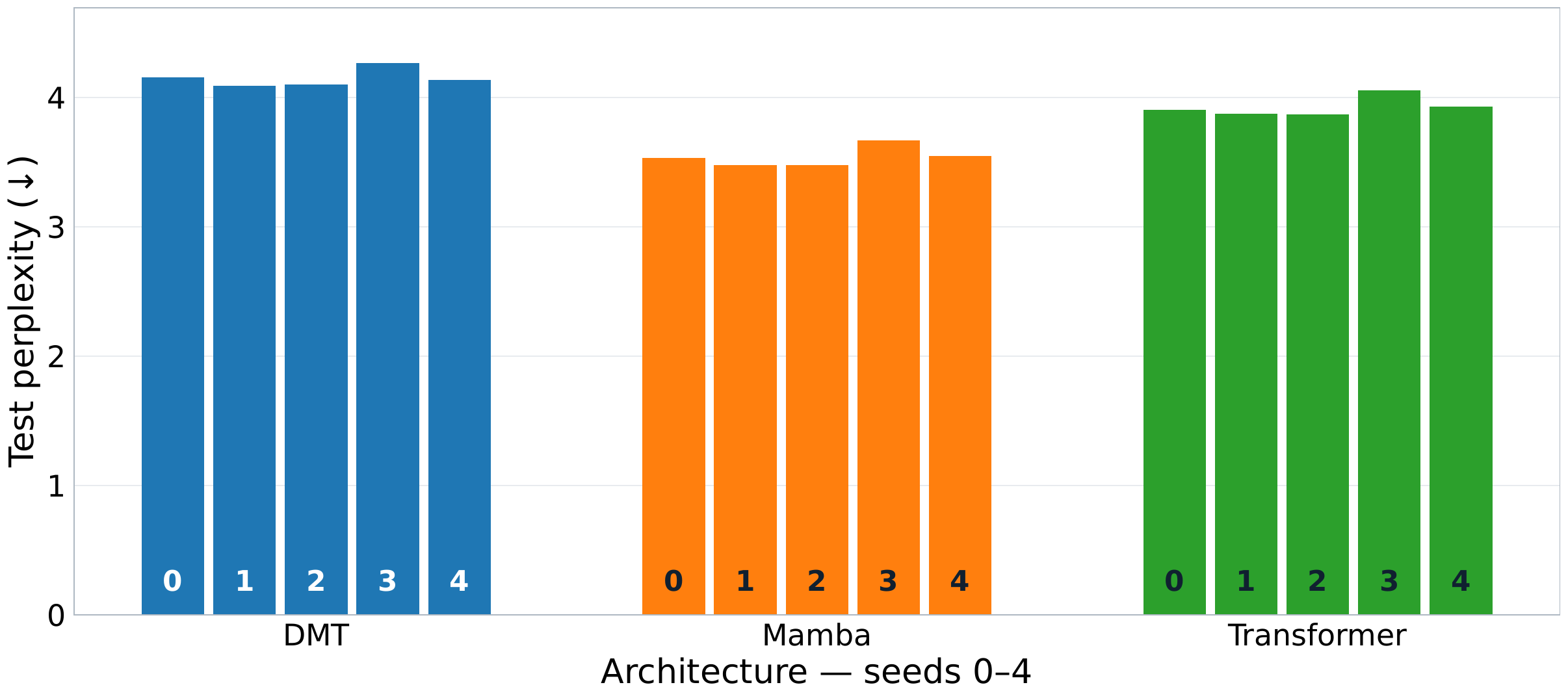}{Test perplexity obtained for each seed from 0 to 4, for the learning rate/weight decay combination selected on validation for each architecture with an arbitrary configuration. These results correspond to the first hyperparameter search phase.}{xp-tiny-seeds}

\xpfig{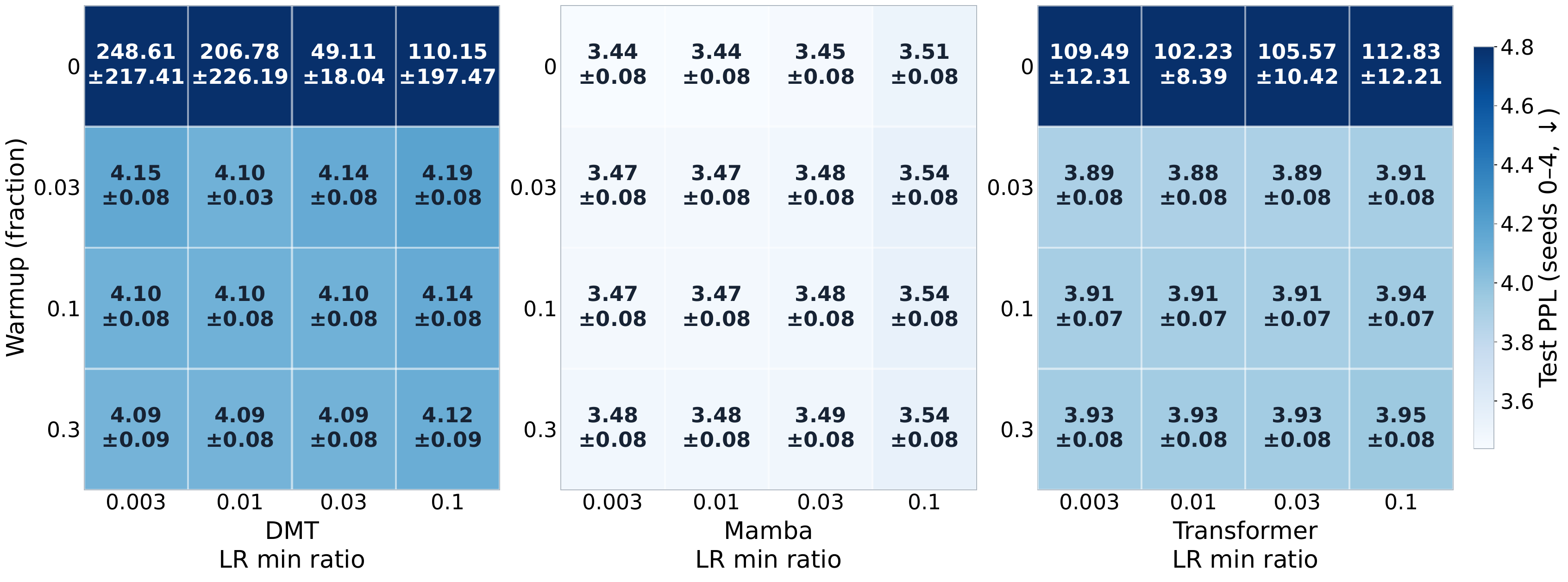}{Mean test perplexity and standard deviation as a function of warmup and \texttt{min\_lr\_ratio}, with learning rate and weight decay fixed.}{xp-tiny-warmup}

\xpfig{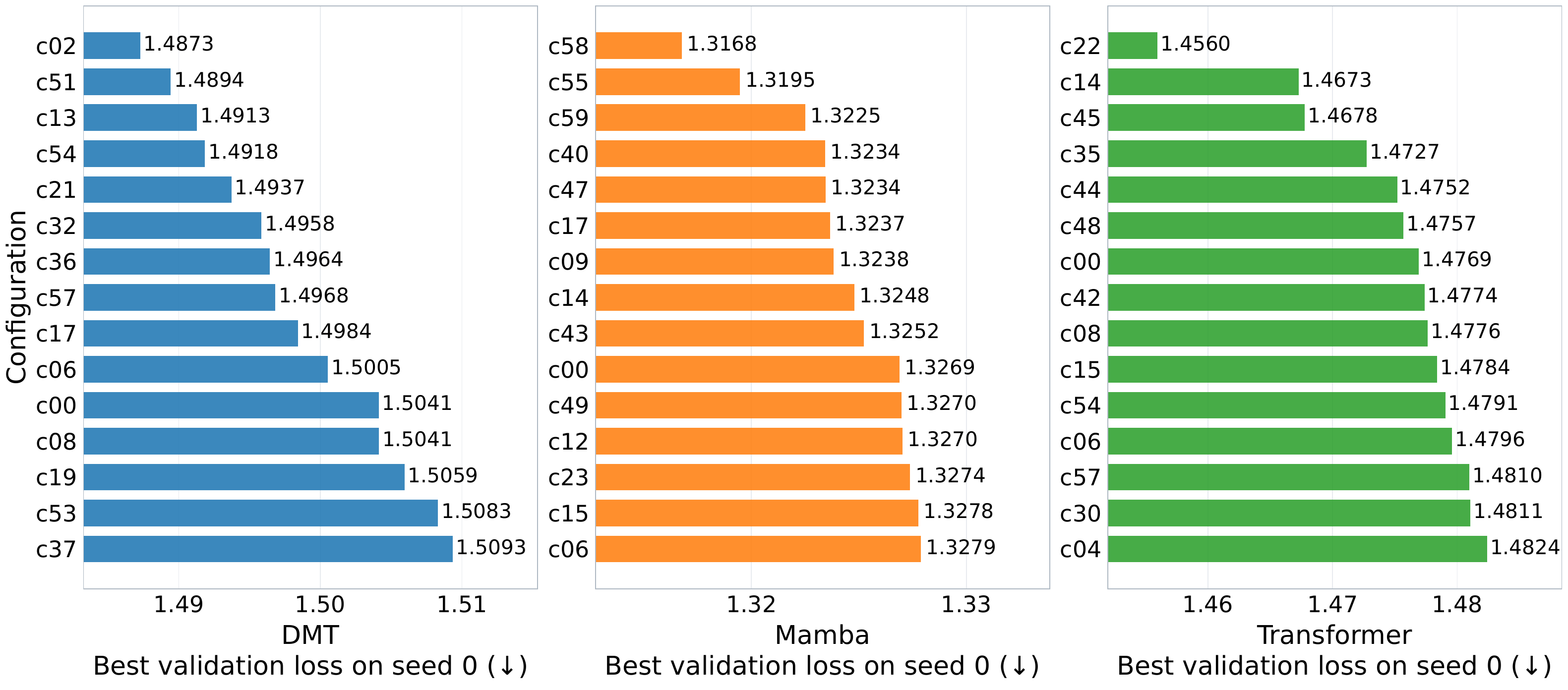}{Validation loss of the top 15 configurations of each architecture after five epochs on seed 0. The $x$-axis differs for each architecture according to its scores.}{xp-tiny-stage1}

\xpfig{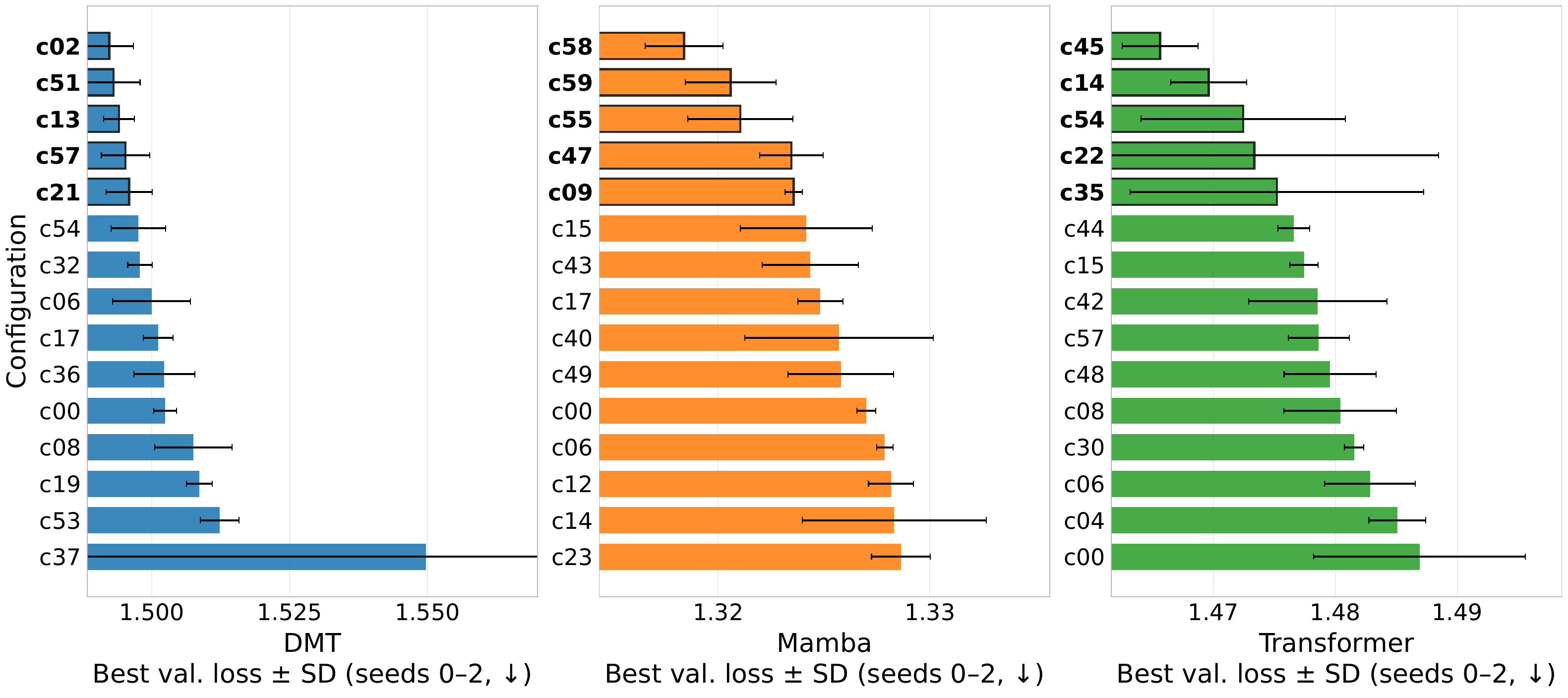}{Mean best validation loss and standard deviation over three seeds, for the 15 configurations after five epochs. Configurations in bold are the five selected configurations for each architecture.}{xp-tiny-stage2}

\xpfig{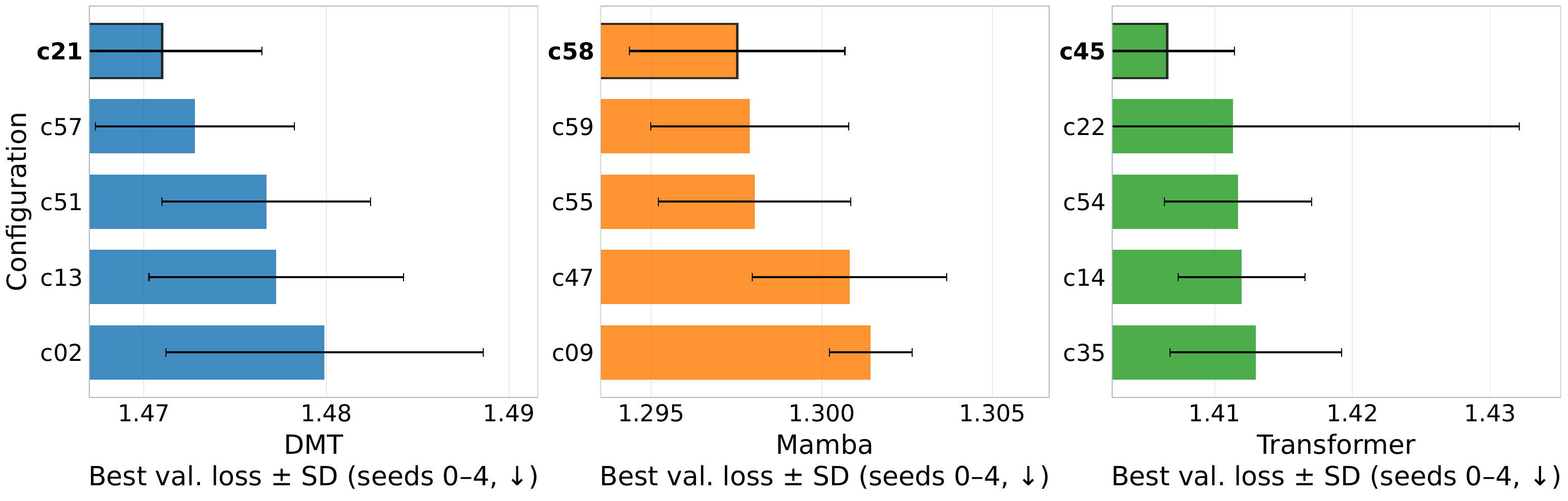}{Ranking of the 5 final configurations for each architecture according to their best mean validation loss, with standard deviation over seeds 0 to 4.}{xp-tiny-stage3-val}

\xpfig{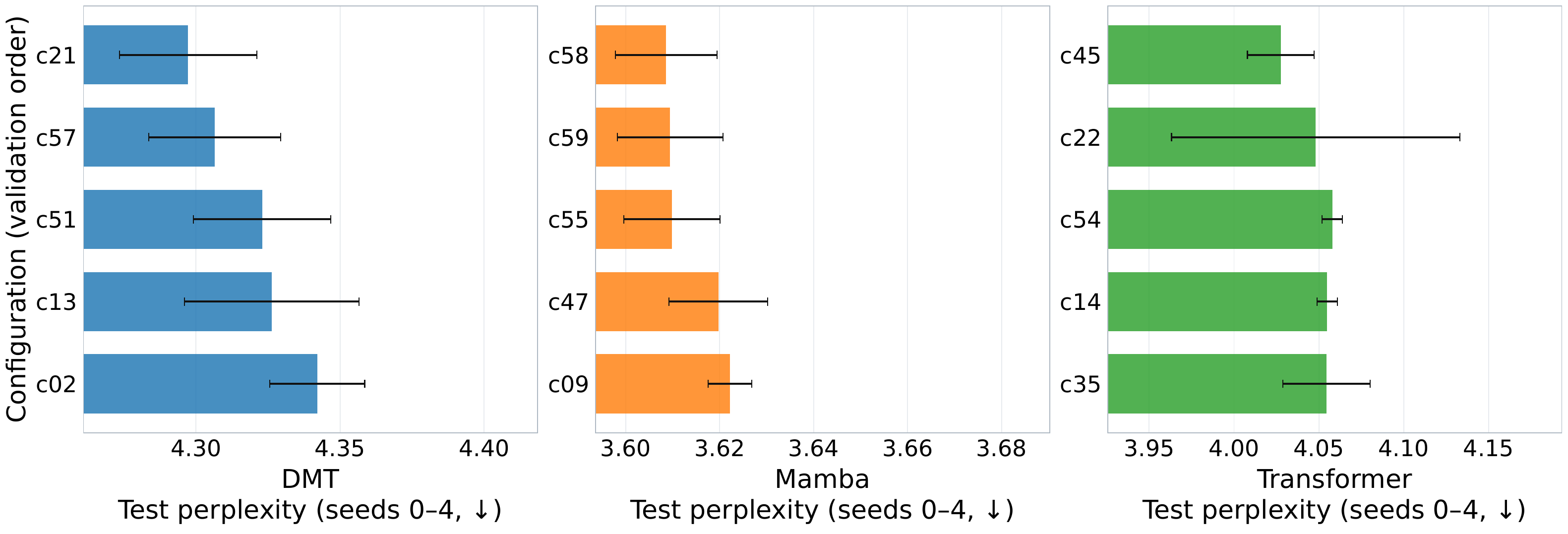}{Mean test perplexity and standard deviation of the final configurations, presented in the order established on validation. Only completed training runs are included; some configurations therefore have fewer than five test seeds. The axes are zoomed separately.}{xp-tiny-stage3-test}

\subsection{OpenWebText}
\label{app:results-owt}

\xpfig{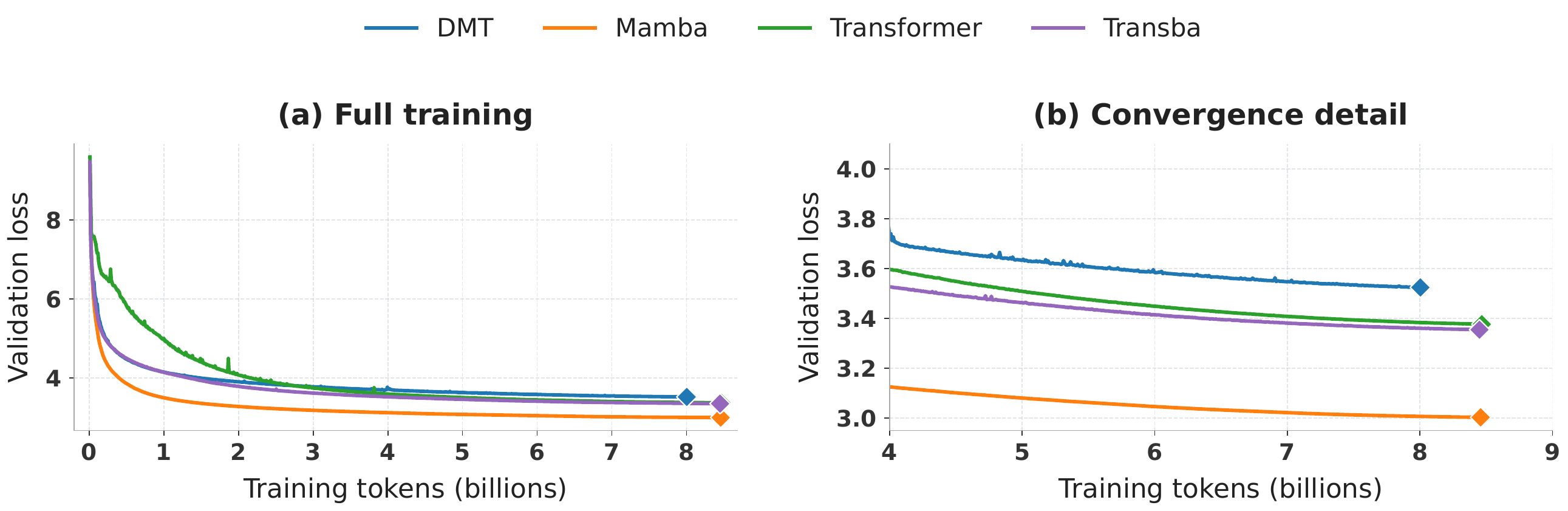}{Evolution of validation loss on OpenWebText as a function of billions of tokens processed. The left plot (a) shows complete training and the right plot (b) a zoom on the end of training. These runs use a single seed. Due to computational resources, DMT stops earlier than the others (see Experimental protocol, Section~\ref{sec:owt}).}{xp-owt-curves}


\section{Configurations explored on TinyStories}
\label{app:tiny-configurations}

We defined 60 configurations per architecture, with a budget between 9 and 11 million parameters. The tables below report the dimensions and parameter counts recorded in the training logs. Identifiers \texttt{c00} to \texttt{c59} correspond to those used in the figures and are specific to each architecture.

\paragraph{Dimensions and common settings.}
$D$ denotes the representation dimension and $L$ the number of blocks. For DMT, $L$ counts decoder blocks after a single memory encoder; $n_{\mathrm{heads}}$ is the number of attention heads, $M$ the number of memory units and $N$ the complex dimension of their state. Convolution width is fixed at $K=4$. For Mamba, the search varies $D$ and $L$, with a state dimension of 16, an internal expansion of 2 and a convolution width of 4. For Transformer, it varies $D$, $L$ and $n_{\mathrm{heads}}$. All configurations use a vocabulary of 2\,048 tokens, a context of 256 tokens and a configured dropout of 0.1.

\begin{table}[H]
\centering
\small
\begin{tabular}{lrrrrr}
  \toprule
  Architecture & $D$ & $L$ & $n_{\mathrm{heads}}$ & $M$ & $N$ \\
  \midrule
  DMT & 160--352 & 4--20 & 2--11 & 2--16 & 64--512 \\
  Mamba & 192--608 & 4--36 & -- & -- & -- \\
  Transformer & 144--448 & 4--36 & 3--28 & -- & -- \\
  \bottomrule
\end{tabular}
\caption{Range of dimensions in the 60 configurations defined per architecture. The bounds do not form a Cartesian grid: only the combinations detailed in the following tables are used.}
\end{table}

\paragraph{Selection stages.}
The \textit{Selection} column indicates the path of each configuration: \texttt{I} denotes a configuration not retained after the initial evaluation on seed 0; \texttt{II}, a configuration shortlisted among the 15 for intermediate repetitions; \texttt{III}, a configuration retained among the five finalists. The configuration finally selected based on mean validation loss is shown in bold and marked \texttt{III*}. Finalists are evaluated on seeds 0 to 4. The marks describe the selection performed, not the presence of all planned runs.

For DMT, \texttt{c21} was retained for the final phase despite its missing intermediate runs on seeds 1 and 2; its selection therefore does not result from a complete ranking over three seeds. Configuration \texttt{c49}, marked \texttt{--}, has no training log: its dimensions and parameter count come only from the configuration registry.

\clearpage
\subsection{DMT}
\begingroup
\footnotesize
\setlength{\tabcolsep}{5pt}
\renewcommand{\arraystretch}{0.90}
\begin{longtable}{lrrrrrrc}
  \toprule
  
Identifier & $D$ & $L$ & $n_{\mathrm{heads}}$ & $M$ & $N$ & Parameters & Selection \\
\midrule
\endfirsthead
\multicolumn{8}{c}{DMT configurations (continued)}\\
\toprule
Identifier & $D$ & $L$ & $n_{\mathrm{heads}}$ & $M$ & $N$ & Parameters & Selection \\
\midrule
\endhead
\bottomrule
\endfoot
\texttt{c00} & 256 & 8 & 8 & 8 & 256 & 10\,509\,064 & \texttt{II} \\
\texttt{c01} & 160 & 20 & 2 & 8 & 512 & 9\,855\,208 & \texttt{I} \\
\texttt{c02} & 192 & 18 & 3 & 16 & 64 & 10\,745\,168 & \texttt{III} \\
\texttt{c03} & 192 & 4 & 3 & 16 & 512 & 10\,066\,000 & \texttt{I} \\
\texttt{c04} & 192 & 20 & 6 & 2 & 64 & 9\,892\,418 & \texttt{I} \\
\texttt{c05} & 352 & 4 & 11 & 2 & 512 & 9\,338\,786 & \texttt{I} \\
\texttt{c06} & 320 & 4 & 5 & 16 & 64 & 10\,851\,792 & \texttt{II} \\
\texttt{c07} & 256 & 10 & 4 & 2 & 512 & 10\,236\,674 & \texttt{I} \\
\texttt{c08} & 352 & 4 & 11 & 8 & 64 & 10\,116\,392 & \texttt{II} \\
\texttt{c09} & 288 & 8 & 4 & 2 & 64 & 9\,632\,738 & \texttt{I} \\
\texttt{c10} & 160 & 16 & 5 & 16 & 384 & 10\,357\,232 & \texttt{I} \\
\texttt{c11} & 192 & 20 & 6 & 2 & 384 & 10\,385\,218 & \texttt{I} \\
\texttt{c12} & 192 & 14 & 3 & 8 & 256 & 9\,163\,592 & \texttt{I} \\
\texttt{c13} & 224 & 10 & 7 & 16 & 96 & 9\,952\,304 & \texttt{III} \\
\texttt{c14} & 320 & 4 & 5 & 6 & 384 & 10\,425\,542 & \texttt{I} \\
\texttt{c15} & 192 & 12 & 6 & 8 & 512 & 9\,855\,048 & \texttt{I} \\
\texttt{c16} & 320 & 6 & 10 & 2 & 256 & 9\,760\,194 & \texttt{I} \\
\texttt{c17} & 224 & 14 & 7 & 8 & 64 & 10\,629\,416 & \texttt{II} \\
\texttt{c18} & 256 & 6 & 8 & 6 & 512 & 9\,720\,582 & \texttt{I} \\
\texttt{c19} & 256 & 6 & 4 & 12 & 192 & 9\,728\,268 & \texttt{II} \\
\texttt{c20} & 224 & 4 & 7 & 16 & 384 & 10\,474\,928 & \texttt{I} \\
\textbf{\texttt{c21}} & \textbf{320} & \textbf{4} & \textbf{10} & \textbf{16} & \textbf{64} & \textbf{10\,851\,792} & \textbf{\texttt{III*}} \\
\texttt{c22} & 192 & 10 & 3 & 16 & 256 & 9\,568\,592 & \texttt{I} \\
\texttt{c23} & 192 & 20 & 3 & 8 & 64 & 10\,637\,384 & \texttt{I} \\
\texttt{c24} & 224 & 14 & 7 & 2 & 256 & 10\,019\,234 & \texttt{I} \\
\texttt{c25} & 288 & 8 & 9 & 4 & 64 & 10\,115\,684 & \texttt{I} \\
\texttt{c26} & 160 & 20 & 2 & 12 & 256 & 9\,406\,444 & \texttt{I} \\
\texttt{c27} & 160 & 12 & 2 & 16 & 512 & 10\,441\,968 & \texttt{I} \\
\texttt{c28} & 320 & 4 & 5 & 8 & 128 & 9\,198\,024 & \texttt{I} \\
\texttt{c29} & 256 & 8 & 4 & 6 & 256 & 9\,718\,534 & \texttt{I} \\
\texttt{c30} & 256 & 4 & 4 & 8 & 512 & 9\,462\,536 & \texttt{I} \\
\texttt{c31} & 256 & 4 & 8 & 16 & 192 & 9\,472\,784 & \texttt{I} \\
\texttt{c32} & 192 & 16 & 6 & 12 & 192 & 10\,545\,740 & \texttt{II} \\
\texttt{c33} & 192 & 18 & 3 & 2 & 512 & 9\,696\,834 & \texttt{I} \\
\texttt{c34} & 192 & 20 & 3 & 4 & 256 & 10\,732\,100 & \texttt{I} \\
\texttt{c35} & 224 & 14 & 7 & 2 & 512 & 10\,479\,010 & \texttt{I} \\
\texttt{c36} & 256 & 10 & 4 & 8 & 64 & 10\,507\,016 & \texttt{II} \\
\texttt{c37} & 320 & 4 & 10 & 8 & 256 & 10\,510\,792 & \texttt{II} \\
\texttt{c38} & 256 & 10 & 8 & 2 & 384 & 9\,974\,018 & \texttt{I} \\
\texttt{c39} & 192 & 6 & 6 & 12 & 512 & 9\,075\,020 & \texttt{I} \\
\texttt{c40} & 320 & 6 & 5 & 2 & 256 & 9\,760\,194 & \texttt{I} \\
\texttt{c41} & 224 & 14 & 7 & 2 & 64 & 9\,674\,402 & \texttt{I} \\
\texttt{c42} & 160 & 20 & 5 & 8 & 384 & 9\,197\,800 & \texttt{I} \\
\texttt{c43} & 320 & 6 & 5 & 2 & 512 & 10\,416\,578 & \texttt{I} \\
\texttt{c44} & 192 & 18 & 6 & 6 & 192 & 10\,094\,918 & \texttt{I} \\
\texttt{c45} & 160 & 20 & 5 & 16 & 192 & 9\,615\,088 & \texttt{I} \\
\texttt{c46} & 192 & 12 & 3 & 12 & 384 & 10\,548\,812 & \texttt{I} \\
\texttt{c47} & 160 & 16 & 5 & 12 & 512 & 10\,148\,588 & \texttt{I} \\
\texttt{c48} & 256 & 4 & 4 & 12 & 384 & 10\,518\,284 & \texttt{I} \\
\texttt{c49} & 256 & 8 & 4 & 16 & 64 & 10\,519\,312 & \texttt{--} \\
\texttt{c50} & 192 & 14 & 3 & 6 & 512 & 9\,802\,310 & \texttt{I} \\
\texttt{c51} & 288 & 6 & 9 & 12 & 64 & 10\,055\,660 & \texttt{III} \\
\texttt{c52} & 192 & 8 & 3 & 12 & 512 & 9\,960\,524 & \texttt{I} \\
\texttt{c53} & 224 & 10 & 7 & 8 & 384 & 10\,518\,056 & \texttt{II} \\
\texttt{c54} & 192 & 16 & 6 & 16 & 64 & 9\,859\,664 & \texttt{II} \\
\texttt{c55} & 192 & 14 & 3 & 12 & 128 & 9\,068\,876 & \texttt{I} \\
\texttt{c56} & 160 & 10 & 5 & 16 & 512 & 9\,826\,928 & \texttt{I} \\
\texttt{c57} & 224 & 10 & 7 & 12 & 192 & 10\,235\,180 & \texttt{III} \\
\texttt{c58} & 192 & 16 & 6 & 6 & 384 & 10\,096\,454 & \texttt{I} \\
\texttt{c59} & 288 & 8 & 9 & 2 & 512 & 10\,666\,722 & \texttt{I} \\\\
\caption{DMT configurations in the TinyStories search. Parameter counts include input and output projections.}
\end{longtable}
\endgroup

\clearpage
\subsection{Mamba}
\begingroup
\footnotesize
\setlength{\tabcolsep}{5pt}
\renewcommand{\arraystretch}{0.90}
\begin{longtable}{lrrrc}
\toprule
Identifier & $D$ & $L$ & Parameters & Selection \\
\midrule
\endfirsthead
\multicolumn{5}{c}{Mamba configurations (continued)}\\
\toprule
Identifier & $D$ & $L$ & Parameters & Selection \\
\midrule
\endhead
\bottomrule
\endfoot
\texttt{c00} & 256 & 22 & 10\,160\,896 & \texttt{II} \\
\texttt{c01} & 608 & 4 & 10\,757\,344 & \texttt{I} \\
\texttt{c02} & 400 & 9 & 10\,219\,200 & \texttt{I} \\
\texttt{c03} & 192 & 36 & 9\,455\,040 & \texttt{I} \\
\texttt{c04} & 304 & 14 & 9\,181\,712 & \texttt{I} \\
\texttt{c05} & 496 & 6 & 10\,572\,240 & \texttt{I} \\
\texttt{c06} & 224 & 29 & 10\,274\,432 & \texttt{II} \\
\texttt{c07} & 352 & 11 & 9\,669\,440 & \texttt{I} \\
\texttt{c08} & 448 & 7 & 10\,046\,848 & \texttt{I} \\
\texttt{c09} & 288 & 18 & 10\,496\,736 & \texttt{III} \\
\texttt{c10} & 208 & 33 & 10\,111\,296 & \texttt{I} \\
\texttt{c11} & 544 & 5 & 10\,664\,576 & \texttt{I} \\
\texttt{c12} & 240 & 25 & 10\,157\,760 & \texttt{II} \\
\texttt{c13} & 576 & 4 & 9\,730\,368 & \texttt{I} \\
\texttt{c14} & 336 & 13 & 10\,346\,112 & \texttt{II} \\
\texttt{c15} & 272 & 20 & 10\,409\,168 & \texttt{II} \\
\texttt{c16} & 224 & 27 & 9\,597\,504 & \texttt{I} \\
\texttt{c17} & 304 & 16 & 10\,404\,400 & \texttt{II} \\
\texttt{c18} & 208 & 31 & 9\,524\,320 & \texttt{I} \\
\texttt{c19} & 416 & 8 & 9\,874\,592 & \texttt{I} \\
\texttt{c20} & 368 & 10 & 9\,626\,512 & \texttt{I} \\
\texttt{c21} & 480 & 6 & 9\,943\,200 & \texttt{I} \\
\texttt{c22} & 528 & 5 & 10\,086\,912 & \texttt{I} \\
\texttt{c23} & 320 & 14 & 10\,112\,960 & \texttt{II} \\
\texttt{c24} & 592 & 4 & 10\,237\,456 & \texttt{I} \\
\texttt{c25} & 288 & 17 & 9\,946\,368 & \texttt{I} \\
\texttt{c26} & 224 & 28 & 9\,935\,968 & \texttt{I} \\
\texttt{c27} & 272 & 19 & 9\,916\,576 & \texttt{I} \\
\texttt{c28} & 240 & 23 & 9\,384\,480 & \texttt{I} \\
\texttt{c29} & 208 & 35 & 10\,698\,272 & \texttt{I} \\
\texttt{c30} & 208 & 32 & 9\,817\,808 & \texttt{I} \\
\texttt{c31} & 304 & 15 & 9\,793\,056 & \texttt{I} \\
\texttt{c32} & 384 & 10 & 10\,429\,056 & \texttt{I} \\
\texttt{c33} & 240 & 24 & 9\,771\,120 & \texttt{I} \\
\texttt{c34} & 512 & 5 & 9\,525\,248 & \texttt{I} \\
\texttt{c35} & 256 & 21 & 9\,722\,880 & \texttt{I} \\
\texttt{c36} & 432 & 8 & 10\,599\,984 & \texttt{I} \\
\texttt{c37} & 336 & 12 & 9\,603\,216 & \texttt{I} \\
\texttt{c38} & 208 & 34 & 10\,404\,784 & \texttt{I} \\
\texttt{c39} & 464 & 6 & 9\,333\,360 & \texttt{I} \\
\texttt{c40} & 352 & 12 & 10\,482\,912 & \texttt{II} \\
\texttt{c41} & 368 & 11 & 10\,513\,760 & \texttt{I} \\
\texttt{c42} & 384 & 9 & 9\,464\,832 & \texttt{I} \\
\texttt{c43} & 240 & 26 & 10\,544\,400 & \texttt{II} \\
\texttt{c44} & 560 & 4 & 9\,236\,080 & \texttt{I} \\
\texttt{c45} & 320 & 13 & 9\,437\,440 & \texttt{I} \\
\texttt{c46} & 272 & 18 & 9\,423\,984 & \texttt{I} \\
\texttt{c47} & 256 & 23 & 10\,598\,912 & \texttt{III} \\
\texttt{c48} & 288 & 16 & 9\,396\,000 & \texttt{I} \\
\texttt{c49} & 224 & 30 & 10\,612\,896 & \texttt{II} \\
\texttt{c50} & 432 & 7 & 9\,385\,632 & \texttt{I} \\
\texttt{c51} & 256 & 20 & 9\,284\,864 & \texttt{I} \\
\texttt{c52} & 464 & 7 & 10\,730\,464 & \texttt{I} \\
\texttt{c53} & 224 & 26 & 9\,259\,040 & \texttt{I} \\
\texttt{c54} & 208 & 30 & 9\,230\,832 & \texttt{I} \\
\texttt{c55} & 320 & 15 & 10\,788\,480 & \texttt{III} \\
\texttt{c56} & 192 & 35 & 9\,203\,328 & \texttt{I} \\
\texttt{c57} & 400 & 8 & 9\,174\,800 & \texttt{I} \\
\textbf{\texttt{c58}} & \textbf{272} & \textbf{21} & \textbf{10\,901\,760} & \textbf{\texttt{III*}} \\
\texttt{c59} & 240 & 27 & 10\,931\,040 & \texttt{III} \\\\
\caption{Mamba configurations in the TinyStories search. Parameter counts include input and output projections.}
\end{longtable}
\endgroup

\clearpage
\subsection{Transformer}
\begingroup
\footnotesize
\setlength{\tabcolsep}{5pt}
\renewcommand{\arraystretch}{0.90}
\begin{longtable}{lrrrrc}
\toprule
Identifier & $D$ & $L$ & $n_{\mathrm{heads}}$ & Parameters & Selection \\
\midrule
\endfirsthead
\multicolumn{6}{c}{Transformer configurations (continued)}\\
\toprule
Identifier & $D$ & $L$ & $n_{\mathrm{heads}}$ & Parameters & Selection \\
\midrule
\endhead
\bottomrule
\endfoot
\texttt{c00} & 256 & 12 & 8 & 10\,033\,408 & \texttt{II} \\
\texttt{c01} & 448 & 4 & 28 & 10\,670\,016 & \texttt{I} \\
\texttt{c02} & 144 & 36 & 9 & 9\,300\,240 & \texttt{I} \\
\texttt{c03} & 448 & 4 & 7 & 10\,670\,016 & \texttt{I} \\
\texttt{c04} & 320 & 8 & 20 & 10\,573\,120 & \texttt{II} \\
\texttt{c05} & 192 & 23 & 3 & 10\,625\,856 & \texttt{I} \\
\texttt{c06} & 192 & 21 & 12 & 9\,740\,352 & \texttt{II} \\
\texttt{c07} & 352 & 6 & 11 & 9\,736\,672 & \texttt{I} \\
\texttt{c08} & 256 & 12 & 16 & 10\,033\,408 & \texttt{II} \\
\texttt{c09} & 384 & 5 & 24 & 9\,736\,320 & \texttt{I} \\
\texttt{c10} & 448 & 4 & 14 & 10\,670\,016 & \texttt{I} \\
\texttt{c11} & 320 & 8 & 5 & 10\,573\,120 & \texttt{I} \\
\texttt{c12} & 160 & 30 & 5 & 9\,594\,400 & \texttt{I} \\
\texttt{c13} & 384 & 5 & 6 & 9\,736\,320 & \texttt{I} \\
\texttt{c14} & 176 & 28 & 11 & 10\,823\,472 & \texttt{III} \\
\texttt{c15} & 224 & 16 & 14 & 10\,157\,280 & \texttt{II} \\
\texttt{c16} & 224 & 17 & 7 & 10\,759\,840 & \texttt{I} \\
\texttt{c17} & 256 & 12 & 4 & 10\,033\,408 & \texttt{I} \\
\texttt{c18} & 160 & 32 & 10 & 10\,209\,440 & \texttt{I} \\
\texttt{c19} & 288 & 9 & 9 & 9\,626\,976 & \texttt{I} \\
\texttt{c20} & 352 & 6 & 22 & 9\,736\,672 & \texttt{I} \\
\texttt{c21} & 192 & 21 & 6 & 9\,740\,352 & \texttt{I} \\
\texttt{c22} & 288 & 10 & 18 & 10\,622\,880 & \texttt{III} \\
\texttt{c23} & 384 & 5 & 12 & 9\,736\,320 & \texttt{I} \\
\texttt{c24} & 416 & 4 & 26 & 9\,268\,896 & \texttt{I} \\
\texttt{c25} & 320 & 8 & 10 & 10\,573\,120 & \texttt{I} \\
\texttt{c26} & 160 & 34 & 5 & 10\,824\,480 & \texttt{I} \\
\texttt{c27} & 176 & 24 & 11 & 9\,335\,216 & \texttt{I} \\
\texttt{c28} & 416 & 4 & 13 & 9\,268\,896 & \texttt{I} \\
\texttt{c29} & 208 & 18 & 13 & 9\,831\,952 & \texttt{I} \\
\texttt{c30} & 240 & 14 & 15 & 10\,236\,720 & \texttt{II} \\
\texttt{c31} & 192 & 20 & 3 & 9\,297\,600 & \texttt{I} \\
\texttt{c32} & 432 & 4 & 27 & 9\,957\,168 & \texttt{I} \\
\texttt{c33} & 336 & 7 & 21 & 10\,262\,448 & \texttt{I} \\
\texttt{c34} & 176 & 26 & 11 & 10\,079\,344 & \texttt{I} \\
\texttt{c35} & 272 & 11 & 17 & 10\,398\,832 & \texttt{III} \\
\texttt{c36} & 160 & 32 & 5 & 10\,209\,440 & \texttt{I} \\
\texttt{c37} & 304 & 8 & 19 & 9\,577\,520 & \texttt{I} \\
\texttt{c38} & 160 & 30 & 10 & 9\,594\,400 & \texttt{I} \\
\texttt{c39} & 400 & 5 & 25 & 10\,526\,000 & \texttt{I} \\
\texttt{c40} & 224 & 15 & 7 & 9\,554\,720 & \texttt{I} \\
\texttt{c41} & 368 & 6 & 23 & 10\,603\,184 & \texttt{I} \\
\texttt{c42} & 192 & 23 & 12 & 10\,625\,856 & \texttt{II} \\
\texttt{c43} & 192 & 23 & 6 & 10\,625\,856 & \texttt{I} \\
\texttt{c44} & 160 & 34 & 10 & 10\,824\,480 & \texttt{II} \\
\textbf{\texttt{c45}} & \textbf{208} & \textbf{20} & \textbf{13} & \textbf{10\,871\,120} & \textbf{\texttt{III*}} \\
\texttt{c46} & 160 & 31 & 10 & 9\,901\,920 & \texttt{I} \\
\texttt{c47} & 160 & 31 & 5 & 9\,901\,920 & \texttt{I} \\
\texttt{c48} & 224 & 16 & 7 & 10\,157\,280 & \texttt{II} \\
\texttt{c49} & 192 & 22 & 12 & 10\,183\,104 & \texttt{I} \\
\texttt{c50} & 192 & 22 & 6 & 10\,183\,104 & \texttt{I} \\
\texttt{c51} & 192 & 22 & 3 & 10\,183\,104 & \texttt{I} \\
\texttt{c52} & 192 & 21 & 3 & 9\,740\,352 & \texttt{I} \\
\texttt{c53} & 176 & 25 & 11 & 9\,707\,280 & \texttt{I} \\
\texttt{c54} & 208 & 19 & 13 & 10\,351\,536 & \texttt{III} \\
\texttt{c55} & 288 & 9 & 18 & 9\,626\,976 & \texttt{I} \\
\texttt{c56} & 224 & 15 & 14 & 9\,554\,720 & \texttt{I} \\
\texttt{c57} & 176 & 27 & 11 & 10\,451\,408 & \texttt{II} \\
\texttt{c58} & 240 & 13 & 15 & 9\,545\,040 & \texttt{I} \\
\texttt{c59} & 272 & 10 & 17 & 9\,510\,480 & \texttt{I} \\\\
\caption{Transformer configurations in the TinyStories search. Parameter counts include input and output projections.}
\end{longtable}
\endgroup

%% file: generations_owt.tex
\section{Qualitative OpenWebText generations}
\label{app:owt-generations}

\subsection{Part 1: Short continuations}

For each of three fixed English prompts and each OWT model (DMT, Mamba, and Transformer), we generated five continuations at each temperature T = 0.7, 0.8, and 1.0, with top-k = 40 and a maximum of 256 new GPT-2 tokens. We qualitatively selected the best of the 15 continuations independently for each prompt–model pair, yielding nine examples from 135 short generations. We also generated five long continuations per model from the railway-station prompt at T = 0.8, with 2,304 new tokens and EOS masked (15 long generations; 150 generations in total, excluding the pilot). The selected continuations below are reproduced without rewriting; only the EOS marker is omitted.

\paragraph{Coastal ecology.}
\textbf{Prompt.}
\begin{quote}\small\raggedright
The researchers returned to the coast each spring to measure the water, record the wildlife, and compare their findings with earlier surveys. This year, they noticed a change that
\end{quote}

\noindent\textbf{DMT (T = 0.7, draw 04).}
\begin{quote}\small\raggedright
was surprising in the last year.

The researchers also found that the researchers had been able to quantify the impact of the large-scale changes over the past few decades.

"The changes in the water quality in the past are not unexpected," said study co-author R. Stenhouse of the University of California, San Diego. "The findings are just the beginning."

Stenhouse said he would like to see the findings again in the future. "We're seeing that the water changes significantly," he said. "We've had some people take this as a positive feedback. I'm confident it's a positive."

The findings are based on data from the Woods Hole Oceanographic Institution, the University of Hawaii and the University of California, Berkeley.

Copyright 2016 The Associated Press. All rights reserved. This material may not be published, broadcast, rewritten, or redistributed.
\end{quote}

\noindent\textbf{Mamba (T = 0.8, draw 02).}
\begin{quote}\small\raggedright
seemed to go beyond the usual pattern of years of steady increases. There was a dramatic increase in the number of females in the estuaries, and a dramatic drop in the number of males in the estuaries toward the end of the year.

“It’s quite dramatic,” says Dr. Linn, “but it’s not dramatic enough to be cause for alarm.”

Dr. Linn says the scientists have tried many different approaches to understand the estuaries, but at this point the researchers are still in the early stages of development. What they expect to get in the next year is a series of new studies, and the scientists hope to publish a detailed description of what they see and what they learn in the coming years. (They plan to share the findings with the public in the coming months). The findings will probably be published by the end of 2016.

“I do think that, for the sake of time, we have to start putting in more money,” says Dr. Linn. “I don’t think we will be able to achieve that at this stage.”

The findings of the study have implications for the future of human-
\end{quote}

\noindent\textbf{Transformer (T = 0.8, draw 05).}
\begin{quote}\small\raggedright
is not likely in the case of more recent species of birds, which include birds.

“We’re hoping that the current data from the survey will show this is a really interesting phenomenon,” said Stryker. “You can measure how many different types of birds, including those that would be more likely to be a great bird, will be different.”

The team is hoping to demonstrate the diversity of birds in the future as well.

The scientists will be able to conduct the next research phase by analyzing their birds’ genetic material, which will allow them to create a more detailed set of genetic and anatomical patterns for their birds.

The team has previously published two new data-sets for birds that have been published previously.
\end{quote}

\paragraph{Railway station and public library.}
\textbf{Prompt.}
\begin{quote}\small\raggedright
The town council has approved a plan to turn the old railway station into a public library. Supporters say the project will bring new life to the neighborhood, while critics worry that
\end{quote}

\noindent\textbf{DMT (T = 0.7, draw 04).}
\begin{quote}\small\raggedright
it could lead to a flood of people.

A city hall employee expressed disappointment at the construction project and the project's construction.

"It's an issue of time, the city has to deal with the fact that they're making a lot of money off the project," said Tim Boudreaux, the city's director of planning.

The city has asked the public to submit a proposal to the mayor that would include a public hearing on the project.

"I think that's a bad thing," said Boudreaux. "The city needs to make a decision on what the project should be to the public, and we'll do that."

The city's plans are for a site for the Boudreaux Center, a new public plaza that would be part of the City Council's redevelopment planning committee. The commission will be responsible for the project, and the city will seek to create a permanent design that will allow the public to have access to the site to be approved.

Boudreaux said he's "not sure" about the plan, but he believes the city is interested in looking at that option.

"There is a possibility that the City of Boudreaux might want to go somewhere
\end{quote}

\noindent\textbf{Mamba (T = 1.0, draw 04).}
\begin{quote}\small\raggedright
it will open up access to the streets. There will not be a park under the new building.

"This is not a place you go to school to go on the bus line," said council Member Tom McMorrow, of the East Village. "This is something that is a huge part of our city."

Council member Mark Levine, who represents East Village, said he supports the project but worries about what would happen to the neighborhood after the project is complete.

"It's going to be the most expensive building and the least amount of activity and development, so the community is going to have to be really smart about what we do here and how we use that space," Levine said.

Levine and McMorrow disagree on how best to develop the project.

"When you talk about redeveloping and improving the housing, for them to do that, that's not necessarily an easy decision," McMorrow said.

Levine said residents should not have to ask whether their neighborhood needs a new library or what it would be like.

"A lot of these issues really are cultural, not political," he said.

Levine said there's plenty of cultural and historical value to be conserved at this point. Residents
\end{quote}

\clearpage

\noindent\textbf{Transformer (T = 0.8, draw 03).}
\begin{quote}\small\raggedright
the station will be used as a public library in downtown.

Supporters say the station will encourage more people to shop online and have a chance to live on the outskirts of downtown.

But they say the station will make it to the community center, in the hope that the community will make it easier for them to use existing sidewalks and other buildings for recreation.

"It's a wonderful example of how we can make this community better," said the chief of the nonprofit Denver Transit Alliance.

"I want to work with the community on the plan. We want to make the city better."

Read or Share this story: http://dmreg.co/2b4yQ2X
\end{quote}

\paragraph{Mara and the brass key.}
\textbf{Prompt.}
\begin{quote}\small\raggedright
Mara found a small brass key beneath the floorboards of her grandfather's house. On the back of the envelope beside it, someone had written,
\end{quote}

\noindent\textbf{DMT (T = 0.7, draw 04).}
\begin{quote}\small\raggedright
"I think I need to be the one who's in charge."

The housekeeper had never been able to find a way to get around that house through a single door, but it was safe from the house's owner.

"I'll give you a clue," said Mura. "I'm very sorry that it would have taken a couple of days, but I understand what you're doing."

Mura said the house was not a garage but rather a house with a basement. It was a small house, but it was in the same neighborhood as the house.

"I could tell that the house had a lot of things, but that was the only thing that was ever built. I'd never been to a house before."

"I'm a builder now," Mura said, "and I've got a lot of people who are just coming out to the house that I love. It's the best living thing you can have, and I'll have a lot of people who are going to want to work with me. There's a lot of people that have been around for years that've gone through the house."

Mura's house was built in 1875, the third-floor house in the district
\end{quote}

\noindent\textbf{Mamba (T = 0.7, draw 05).}
\begin{quote}\small\raggedright
"Thank you."

"I was just as surprised as everyone else when I found it in the mailbox," she said.

"It was right there in the mailbox, in the box. It was the most precious thing. I don't think anybody has ever seen anything like that in my life."

Mara said it was a nice surprise that she and her husband were able to be in the same room together.

"I'm very thankful for the support, the community," she said. "I'm really grateful."
\end{quote}

\noindent\textbf{Transformer (T = 1.0, draw 03).}
\begin{quote}\small\raggedright
"We won't have a secret plan. I'm not allowed to talk about it yet for now…But we could do it."

She sat down with her parents, their children, their kids, and a few others. To see more of her parents as their kids were in the house, they made her feel guilty as they made her feel guilty. She told her mother how her family went to check the boxes when she was young. They all said they had a list of 10,000 boxes in one drawer and the number was closer to 200 in one drawer. She even kept herself safe, as it turned out the only way to escape from the house was to enter the house.

The children had to go inside inside the house, so they were left alone to escape from the house. One of the children tried to grab a bag of paper and said he wouldn't talk to her since it was so empty. He said that he didn't want anyone to take her down. He tried to steal the bag from him and he even tried to break a window. The family then put a hole in the wall that had to be filled with clothes. There was a knife in the basement of the house on top of it.

And that's the
\end{quote}

\subsection{Part 2: Long continuations}

Using the railway-station prompt above, we selected the best complete 2,304-token continuation per model from five draws at T = 0.8, with EOS masked.

\noindent\textbf{DMT (T = 0.8, draw 03, 2,304 new tokens).}
\begin{quote}\small\raggedright
\def\UrlBreaks{\do\/\do-\do\.\do\_}
the town’s main public buildings will be redeveloped.

“It’s a great idea,” said Coun. Tim Poulter, who was the city’s planning director at the time of the project. “It’s a great idea, but it’s so important to us that the plan is to go into the city’s infrastructure, and the city has to become a place where people can build their own place.”

Article Continued Below

The site is the brainchild of Robert Poulter, who said he helped build a community hall in North Berwyn to help raise money for the site. It’s been helping to organize a public outreach campaign to help fund the site, which he hopes will help make the site possible in a big way. Poulter said the \$5,000 goal is a step in the right direction, but will also provide the necessary funding to get the site on track to completion. The site is open from April 1, 2019, to the end of the year.

“With the potential to transform a community, [the site] is a very big opportunity for us,” he said. “It’s a challenge. It’s a challenge to put that community first.”

Poulter said he would take the site up by September, but the site would be located next to the site and will be open before that. He said the site is open to residents, tourists and visitors of the site, and is also open to people to join the site, where they could connect to the site’s sites.

“I think if the site is going to grow on us, we can’t wait to get the site ready,” he said. “We really need a site that would let people know about it.”

Saunders said the site should be able to work with the community as well as outside the city.

“We know that there is a lot of interest in this site because it has a lot of people who have been working for years,” he said. “We’ve been talking about it for a very long time. We’re going to keep doing that.”

Staff Sgt. James Pyle, a partner at the B.C. Liberals Foundation, said the site will “truly get out and try to build a new site.”

“We’re not going to rush to the site anymore. We’re going to give people a place to learn to walk,” Pyle said. “We have a lot of work to do, and we’re not going to get the time to go to that site.”

Contact Anthony D. Anderson at alex@jjournal.ca.

Read or Share this story: \url{https://www.shafayette-bay.com/story/life/2017/05/08/shannon-new-wars-will-build-a-home-a-front-and-the-home-of-the-the-community-and-the-snow/1054093/}

Email to the site: [email protected] (Sun-Times-Google-2-01-09-30-04-26-09-00-00) Hp.

Like us on Facebook and Twitter at \url{www.facebook.com/hp-praxe.com.}

Please adhere to our commenting policy to avoid being banned. As a privately owned website, we reserve the right to remove any comment and ban any user at any time.Comments that contain spam, advertising, vulgarity, threats of violence and death, racism, anti-Semitism, or personal or abusive attacks on other users may be removed and result in a ban.-- Follow these instructions on registering...

Please enable JavaScript to view the comments powered by Disqus.

Disqus

The full story is in our history, and for more about the incident, or for any further information on our website.

Read the full story.

Actions speak louder than words and sentences, and are always free to express their opinions. The site is open to all users, including the community, and is available to everyone in the community.

To read an article at the bottom of this page, go to the comments section below. We will update this page once we have a response. Please read the following content:

"We have received many inquiries about our content, which is being treated unfairly, and we have a very close relationship with the site. Our community is fully informed of our content and is working closely with us to complete the site.

"We're aware of the comments and the use of the site for discussion. We do not condone any part of this content. We are always looking to educate and enlighten our readers.

"We have a reputation for being open and honest about all of these issues - and now we are doing everything we can to make sure that the reader is not in the right place."

The site also includes the discussion section for users to comment on in the page itself, and the discussion of topics.

About the site

The site provides a forum where you can write about issues you care about and have a good laugh. As someone who uses the site, I hope that you will feel welcome to participate in it.

To learn more about the forum, please see the project page.

Questions can follow.

Please email us with questions.

Thank you!

Riot Monsoon

The site is free, and the site is open to the public.

Please click on the link below to visit the site.

If you’d like the site to survive, you can find more information here. We’ll be sending out the link to the site at the same time as we search.

If you have any further questions, please feel free to contact us by email (dot.com) with any relevant information you may have.

We will be hosting the site at \url{www.davea.com} with an official announcement. The site is not currently listed as a site that has been closed.

We have decided to continue to support the site as we continue to work with our community.

We are also looking for another new site to host and publish your own content.

We are currently in negotiations with the site’s owners to ensure the site remains open.

We have no intention to destroy any existing website or any other sites that are a part of the site.

If they do not agree, we will discontinue the site, however the site does not affect your existing site.

If the site is compromised or has a history of infringing activity, we will contact you at the site when we have completed the site.

We are in contact with the site owners and will contact them privately.

Update 7/29/13:

Original email address:

Dear reader,

We are glad to inform you that we have secured information and are aware of how to proceed.

The site will be offline for a few days after the site is closed, and an hour after that, we will be back up in the air for a few days. We are planning to return to our site as soon as possible.

Please leave a feedback form, and we shall update the site accordingly.

We have a number of areas to contact you with, and you can post as many as you wish. If you are unable to reach us, we will contact you immediately.

Please contact us if you would like to contact us.

Please note that your contact information has to be removed from the site for any further communication.

Please make sure that no unauthorized access, contact other sites under the same guise.

We are open for business. We are aware of your situation and the possibility of an alternative site.

We will update this site when more information is reported.

Please remember that people can access our site for all information regarding the site.

We have not received additional information regarding the potential site location of the site.

We have not been notified by the site visitor to any site or website.

We will comply with all the appropriate terms and conditions.

We will do everything possible to ensure that the site is safe to enter and that it is safe to do so.

We will not be providing information to any person or entity outside of the site.

We will ask that you provide information about all of the sites that are subject to the same type of threat.

We will not post any comments on the site at the time of any investigation.

We will send out a warning to all sites that they are not affected by this site.

We will only take precautions if there is an online threat of your site being compromised.

We are working on a fix to our site.

We are not sure if this site is located in the United States or if we are in danger of becoming a suspect.

We are unsure if there is an investigation at this time.

We do not know how many people have been harmed or are at the site.

We are investigating whether or not we have any information that may cause a permanent public investigation.

We will be working with the police to review and further ensure that anything we may be able to get through.

We will not be issuing or sending a message to any site or other site regarding the site.

We will continue to monitor the site regularly to ensure that the security measures are properly enforced.

We will continue to work with our law enforcement partners as necessary to ensure that they are protected, protected and protected from any and all threats.

We will not tolerate any harm that could harm anyone or anything.

We will not tolerate or condone any kind of violence or threats of violence directed against our site or our site.

We will continue to monitor and track all sites that we know of, but we will not be hosting any posts in any form or anywhere outside of our site.

We cannot take any measures against anybody that violates our guidelines.

We will not tolerate anyone posting any content that is deemed offensive or provocative.

We will not censor sites that discriminate against one another.

We will continue to work with law enforcement to ensure that we comply with all requests to remove from access to our sites.

We will not be blocking sites that attempt to influence our sites and processes.

We are not blocking or shutting down sites because of these takedowns.

We will not be deterred from using any website as a location or site for such activities.

We will not tolerate or support attempts to interfere with or disrupt sites or sites being blocked or blocked.

We will not block
\end{quote}

\noindent\textbf{Mamba (T = 0.8, draw 04, 2,304 new tokens).}
\begin{quote}\small\raggedright
\def\UrlBreaks{\do\/\do-\do\.\do\_}
the building’s history and historic context will be forgotten.

“This is a unique part of town, and it’s a huge asset to the community,” said former mayor Robert H. Fitch Jr. “This is a city of public buildings. We are going to have a great space.”

The development was approved by the Planning Commission and approved by the Town Council in February.

The project is intended to create a new community hub of about 70 miles, and to help build a community center and entertainment hub that will be used for community outreach, entertainment events and business development.

The town council approved the project in a Nov. 6 vote, with Councilman Nick Wagoner, a Pittsburgh Republican who represents the area, saying the project will bring “a whole new dimension.”

He said the town council’s approval of the project was based on a “best-case scenario” that would allow for more than 20,000 visitors.

The town council had said it would be able to build up to 25,000 square feet of new residences, as well as more than \$1 million in development improvements, but they have not been able to come to a vote.

The council has yet to give its approval.

The project is the latest in a series of development plans in town.

Earlier in the year, the council announced plans for a \$200 million mixed-use development at the intersection of Pennsylvania Avenue and Broadway in downtown Pittsburgh.

The project is the latest of 11 projects that have been proposed for the neighborhood, but it has faced opposition from developers.

A public hearing was held last week on the project, which was pushed in part by residents and city officials, who want the development to be in a more urban neighborhood.

The project will be funded by the state, the city of Pittsburgh, the Pennsylvania Department of Transportation and the Allegheny Regional Commission, according to the city’s website.

In a statement, Mayor Bill Peduto said the project was “part of a larger effort to revitalize the city and attract and retain residents.”

He said the city would also partner with the city to develop a new neighborhood center and to “rebuild the park around the station.”

The town council also has been conducting regular town hall meetings with the developers and the public during the summer.

A public hearing on the project was held at the city hall on Wednesday.

— Reach AJ Green at 523-945-2195, ogreen@delawareonline.com or on Twitter @at AJGreen. Follow us @livescience, Facebook and Google+. Original article on LiveScience.

Related stories

A town hall meeting with a new development in Lancaster

Will Penn Station get a new name?

Penn's new neighborhood center could become tourist attraction

What is Penn Station?

When the Penn Station project was announced in 2011, it was hailed as “the next great thing in town.”

Penn Station remains the nation’s oldest and largest public transit system.

“There’s a great mix of people, a little bit of history,” said Bill Pflueger, a Pittsburgh-based historian and author who has published numerous books on Penn Station. “I think it’s going to be a very important part of the future of Penn Station.”

Pflueger and the other city officials in town have been working on the project for years.

“It was a real challenge,” Pflueger said. “I thought the project had a lot of potential, but I didn’t think it would go forward.”

The project was approved by the town council, and the city approved the project in May.

The project involved moving a number of historic buildings, including a museum, a library, a new library and a library, to the new site.

The project’s completion was delayed by a few years in 2013, when the buildings were finished, and in early 2015, the city decided to abandon the project after a request from the city.

The project was delayed until 2017.

Pflueger said that the project was in the works even as early as February.

The city and the state of Pennsylvania are currently working on the project.

The project will replace the \$1.8 million building that was once home to Penn Station.

The project will include a new building on the western edge of the station, as well as a pedestrian plaza and a restaurant, according to city officials.

A portion of the project will be dedicated to changing the public space of Penn Station.

“We are very excited about the future of Penn Station,” Pflueger said. “It is a place that really lives up to what the city has offered us.”

Penn Station is slated to be finished in 2021.

The project will be overseen by the city of Pittsburgh and the Pennsylvania Department of Transportation.

The project is scheduled to go before the board in late March.

The city has not yet chosen a location for the project.

“I’ve never been to a site in Pennsylvania when it comes to a new development where there is something unique and unique,” said Pflueger. “It’s hard to find a place to go and a location where it is not a great place to have a party.”

The city is also looking for a third location for the project.

"The last plan we looked at was the Pflueger/Pflueger project in the mid-1980s," said Pflueger. "And when we look at it again, we just want to make sure that the site meets both the city requirements and the quality standards that Penn Station has.”

The city's new transportation plan includes a series of new projects to meet the city's needs.

“I hope that it will give us the ability to find a better site for the station as well as a better location for the new project," Pflueger said.

The new station will be at least three stories tall and will have a capacity of 4,000 seats.

The new project will also add an enclosed walkway to the street.

In November, Penn Central Station will be in the spotlight for a new project, which officials are predicting will transform a half-mile stretch of Pennsylvania Avenue.

“This will open up that possibility of an even more attractive option for Penn Station,” said Bill Pflueger, president and CEO of Pennsylvania Economic Ventures, an economic development group.

“This is a very large corridor. It will mean that we are in a very different state than it has ever been without Penn Station.”

Penn Station will be the first station that will be able to accommodate more people.

“It is going to be the first, the most diverse, the largest, the tallest, the least crowded station in Pennsylvania,” Pflueger said.

Penn Station is the largest transit hub in Pennsylvania, and the center of Penn State University.

“It’s something that will help add to our community,” Pflueger said.

That’s exactly what Pflueger and other developers hope.

“We’ve been told that Penn Station will be a community hub,” Pflueger said. “I’m sure it will transform the rest of the community.”

The city has a history of providing a lot of public space for the general public without the need for expensive and complicated public improvements.

“The city and the state of Pennsylvania are the two biggest supporters of this project,” Pflueger said. “If it is the largest public space in the state of Pennsylvania, it would be a tremendous asset for the city.”

The project is expected to get a lot of public input during the summer and fall.

“We could build a project that really is a real game changer for the city,” Pflueger said.

The project is being called “Penn Station for the Heart” and is being financed by the Pennsylvania Department of Transportation, the state of Pennsylvania and the city of Pittsburgh.

The project will cost an estimated \$3.6 million, according to Mayor Jim Pflueger.

“Our community has been very supportive and encouraging to the developer in terms of having local officials come out and say, ‘Hey, we want to see this,’ ” said Pflueger. “We need to show that we can do a workable project.”

While Pflueger said the project will be a success, he is not sure the project will become a major financial success.

“The project won’t be done,” Pflueger said. “It will be something that is going to be a major financial success.”

The project will provide a high-tech, high-value, transportation hub for Penn Station.

“So, we’re going to have this great public hub,” Pflueger said. “We’re going to have this great pedestrian and cycling center.”

The project also will involve the development of a public plaza and plaza.

“It will be a very popular, great public space,” Pflueger said.

He also said the project will also add a new restaurant, a new library, a more expansive library and the first new Amtrak station in the state.

“That’s going to be a big boost to the project from a historical standpoint,” Pflueger said.

The city will have the option of spending an additional \$300,000 to make the project a success.

Penn Station would be a major economic boost for the state, not just for Penn Station.

“I think what’s really important is we are going to get the best possible public transit service,” Pflueger said. “We are going to be able to drive this for years to come.”

The project will cost about \$500,000. For the first, estimated cost of \$1.6 million, the project will be \$5 million.

Penn Station will host the first meeting of the Urban League, a free public policy group founded by Penn State president Richard B. “Buddy” McKeon, as well as a wide range of other groups.

“We’re going to take a seat on the
\end{quote}
\clearpage

\noindent\textbf{Transformer (T = 0.8, draw 05, 2,304 new tokens).}
\begin{quote}\small\raggedright
\def\UrlBreaks{\do\/\do-\do\.\do\_}
it will become too expensive and the lack of funding continues to worsen. "It's just a waste of money," said Councilman David E. Gershman, a retired professor of architecture at Harvard who has worked at the Central Library. "They were very generous and generous people, and that's a good thing for the neighborhood."

The project's success also comes after the city and county governments agreed to pay the developer \$100,000 to build the new subway station at the city's existing station. But critics say it will cost taxpayers as much as \$100,000. The city has been trying to help its aging subway station, which opened in 2006 but has been plagued by lawsuits, and now it is spending \$10 million to build the station.

Critics say it will cost taxpayers as much as \$10,000, but critics say it will cost taxpayers as much as \$10,000, and that's a good thing for the neighborhood; and they would be disappointed if the project is too expensive and the lack of funding continues to worsen.

"We're not going to pay for the aging subway station," said Peter Ostrovsky, a spokesman for the Central Library's board of directors. "We know that if this project is going to be successful, it will be costly. The cost is very high. The city should have spent \$10 million on building the new station and then the city should have spent \$10 million on building the new station."

City officials on Thursday expressed concerns about the project, citing concerns that the city may be moving money to pay for it.

"That's the cost of doing that," said Gershman, who was an officer for the Central Library, which operates in New York City. "It's not the cost of doing that."

The city is reviewing the MTA's \$1 billion cost to fund a \$6 billion transit project. The city will also approve \$12.5-million in funding to fund the construction of a new subway station along the eastern edge of New York City, said Andrew W. Miller, the city's director of economic development.

The city also is reviewing the MTA's \$1 billion cost to fund a \$6 billion transit project along the eastern edge of New York City, said Peter Ostrovsky, a spokesman for the Central Library, which operates in New York City.

"This isn't a project that has been successful, but it's the cost that's going to hit the city, and that's something that's going to be missed," he said.

The city will provide \$75 million in funding for the project, but it will not pay for the project.

"We're not going to pay for the project," said city officials.

The city has said it is considering a feasibility study that would determine whether it can raise the money.

The project has been delayed in part because of delays, but the city is also mulling an alternative route, said David A. Gershman, a professor of architecture at Harvard.

"I think that this project and the MTA are going to have a real positive effect on the city," he said. "It's going to have a positive impact on the city."

The city could use a \$150 million contract to give the project an operating budget of \$18 million. That would be \$9 million to \$20 million.

The city would also need to make \$100 million in public funds, but it still needs \$6 million to get the project to meet its budget by the end of this month.

The city is looking at \$12 billion for the project. The city is offering \$1 million to \$5 million in additional funding for the project, which would require a \$1.3 million appropriation. The city has already contributed \$1.2 million to the project, and has pledged \$1.25 million to the city.

That money could come from the Central Library's \$20 million project, which was not approved by the city.

The mayor has made a point of not giving money to the project. He has not responded to his critics' questions about the project, but the city is considering the construction of the new station by the end of this month, said city spokesman David E. Gershman.

"The City is investing in the MTA's subway system and I'm not optimistic as to whether that's the right thing to do," he said. "The plan is not to give money to the project."

A spokesperson for the Central Library's downtown office said it is not involved in the negotiations with the city.

The office of the city's executive director of public affairs, Marc K. Miller, also said some of its members have criticized the city for failing to respond to the requests of the project during the budget process.

On Sunday, the city and city officials, who are in a special meeting on Sept. 11, asked the city to make the initial \$10 million investment available to the project and a city-wide public library that includes the subway extension and a \$2.5 million "city transportation," according to the mayor's contract.

The city also received \$1.2 million to \$3.5 million in city funding.

"We are very disappointed that the project is not a project that has been successful," said Gershman. "The city will be taking \$10 million in funding to build the station."

There are concerns about the project's future, said the mayor and city officials. But a city-wide funding decision will need to be made in order to meet the initial \$2.2 million in funding to build the station, said CDP chairwoman of the city's planning committee, Mark Thompson.

Thompson, a former city clerk, said the city and city both have a strong interest in the project and said they are not looking at a project as successful as the city's earlier \$1.2 million for transit projects.

The mayor of New York said he is hopeful the City will consider the project on the city's first day of business.

"It's a long road ahead of us, but hopefully we won't have to see it in the future," said Thompson, who is from New York.

On Friday, the city will begin planning the \$15 million project for the Central Library of New York, which was delayed until late last year.

The city has already pledged \$500 million to build the station, which will begin the next phase and will include the transit project on a site called the Central Library, which will be the site of the new station.

The city has also asked the MTA to consider a funding review for the project, which it is asking the city to approve of.

The mayor said he and the city are "reinforcing the MTA's commitment to the transit system."

The city has not made a final decision on the project, which has been scheduled to be completed by late November, but the city is still seeking money.

The last day of business is expected to begin Nov. 14, the mayor said.

The city has set an April 2018 deadline for all projects, including the MTA, to be completed by December 2018.

The MTA has not announced a date until Jan. 1, 2018.

The city has not given the city a timetable for the project, but the city has agreed to postpone the project until it receives a proper review of its plan.

"We're excited to announce the date for all projects that have been completed and that there will be a time when the MTA has come up with a plan for the project," said Mayor Bill de Blasio.

The \$5 million project has not yet been completed, but some members of the city of New York have been trying to get it done, said John K. Burch, president of the city's Department of Transportation.

"I have been very clear in my mind that there are multiple proposals coming up that are going to be the same," said K. Burch.

A \$2.5 million grant from the city will be required to complete the project, said city spokeswoman Stephanie Grubbs. The mayor also said the project would not be used for the city's transit system.

The mayor said no funding will be required.

"I think this is an opportunity for us to have a look at it," said Mayor Bill de Blasio. "I think it's good to have a look at the project, but we've got to have that to begin with."

The mayor said he is "deeply concerned" about the project being delayed.

"This is a very important project for New York and the whole world," he told reporters. "This is an important one."

Councilman Richard M. Fiedler, who is in charge of the project, said he's hopeful "there will be something to sort of get done" between the project and city staff.

"I'm very excited about the project," said Fiedler. "I know it's a long road ahead of us. It's a long road ahead."

The city has not yet responded to the project's request for a grant from the city to work on its projects.

Commissioner Richard T. M. Gersh of the city's planning committee approved the project earlier this week.

The city will receive up to \$25 million in direct funding from the MTA, according to city officials. But M. Gershman said the city will not make a final decision on any projects.

A city spokesman said the project is not a project by the MTA, and the city officials are not expected to discuss the project at this time.

Gershman said it is a "very positive" finding.

"We have to make sure we don't have too much money in the future," he said.

The mayor said the city's transit and transportation programs are separate entities, but they are separate entities.

The city will have \$500 million in funding to build the station, with funding for the project under \$100 million. The mayor said that it is the second-largest project on the city's plans as of this morning.

The mayor, who declined to give details, said the city is looking at a \$5 million project by the end of this month.

The mayor said the city is still trying to get the project completed, but the MTA has to approve of the project for a period of time.

Mayor Michael Bloomberg on Thursday called the project as a "significant step forward."

"This piece of information will make it easier than ever to develop a project for the next 100 years," Bloomberg said.

The council has not yet asked the mayor to approve the project at this time.

The mayor said he hopes the city will be able to provide more funding in the future, and said he expects the MTA to have its own set of projects in place immediately.

"You have to make sure
\end{quote}

\subsection{Part 3: TTFT and tokens per second}

\begin{table}[b]
\centering\small
\setlength{\tabcolsep}{4pt}
\begin{tabular}{lrrrr}
\toprule
 & \multicolumn{2}{c}{Short} & \multicolumn{2}{c}{Long} \\
\cmidrule(lr){2-3}\cmidrule(lr){4-5}
Model & TTFT (ms) & Tokens/s & TTFT (ms) & Tokens/s \\
\midrule
DMT & 12.36 & 217.02 & 14.35 & 208.32 \\
Mamba & 10.75 & 121.60 & 10.94 & 120.88 \\
Transformer & 4.89 & 263.71 & 5.00 & 63.23 \\
\bottomrule
\end{tabular}
\caption{Arithmetic means over all 45 short generations or all five long generations per model, on one NVIDIA GeForce RTX 3090, batch size 1, FP32. TTFT includes prompt processing; checkpoint loading and tokenization are excluded. Decoding rates include sampling. DMT and Mamba retain recurrent states; Transformer recomputes a rolling 1,024-token window after its context limit.}
\label{tab:owt-generation-speed}
\end{table}